\documentclass[]{fairmeta}
\usepackage{times}
\usepackage{latexsym}
\usepackage{times}
\usepackage{latexsym}
\usepackage{amsfonts}       
\usepackage{nicefrac}       
\usepackage{microtype}      
\usepackage{xcolor}         
\usepackage{tabularx}         
\usepackage{graphicx}
\usepackage{tikz}
\usepackage{pgfplots}
\usepackage{verbatim}
\usepackage{amsmath}
\usepackage{subcaption}
\pgfplotsset{compat=1.18}
\usepgfplotslibrary{groupplots}
\usepackage[T1]{fontenc}
\usepackage{booktabs}
\usepackage[utf8]{inputenc}
\usepackage{microtype}
\usepackage{inconsolata}
\usepackage{graphicx}
\usepackage[T1]{fontenc}
\usepackage[utf8]{inputenc}
\usepackage{microtype}
\usepackage{inconsolata}
\usepackage{twemojis}
\usepackage{collcell}

\definecolor{replace}{RGB}{163, 206, 245}
\definecolor{no_replace}{RGB}{220, 250, 255}
\definecolor{replace_alt}{RGB}{245, 206, 163}
\definecolor{no_replace_alt}{RGB}{255, 250, 220}

\newcommand{\methodname}{CoSMoEs}
\title{\textit{\methodname}: Compact Sparse Mixture of Experts}

\author[1]{Patrick Huber}
\author[2, *]{Akshat Shrivastava}
\author[3]{Ernie Chang}
\author[1]{Chinnadhurai Sankar}
\author[1]{Ahmed Aly}
\author[1]{Adithya Sagar}

\affiliation[1]{Meta Reality Labs}
\affiliation[2]{Perceptron}
\affiliation[3]{Meta GenAI}

\contribution[*]{This work was performed while at Meta.}

\abstract{
Sparse Mixture of Expert (MoE) models are popular foundational architectures at large scale, however, under-explored at smaller sizes. Here, we show how to enable Compact Sparse Mixture of Experts (\methodname) for on-device inference. Specifically, we tackle the three main on-device dimensions: Quality, Memory and Latency. Along the quality axis, we show that in a fair evaluation (removing confounding factors) MoE architectures outperform FLOP-aligned dense models at on-device scale. We introduce weight-decomposed experts, further improving the MoE model performance. Regarding model memory and latency, we significantly improve model offloading efficiency and, in turn, reduce model inference latency.}

\date{\today}
\correspondence{Patrick Huber at \email{patrickhuber@meta.com} and Adithya Sagar at \email{adithyasagar@meta.com}}

\metadata[Code]{Model code and checkpoints will be released soon!}

\begin{document}

\maketitle

\begin{figure}[t]
    \centering
    \resizebox{0.85\linewidth}{!}{
    \includegraphics{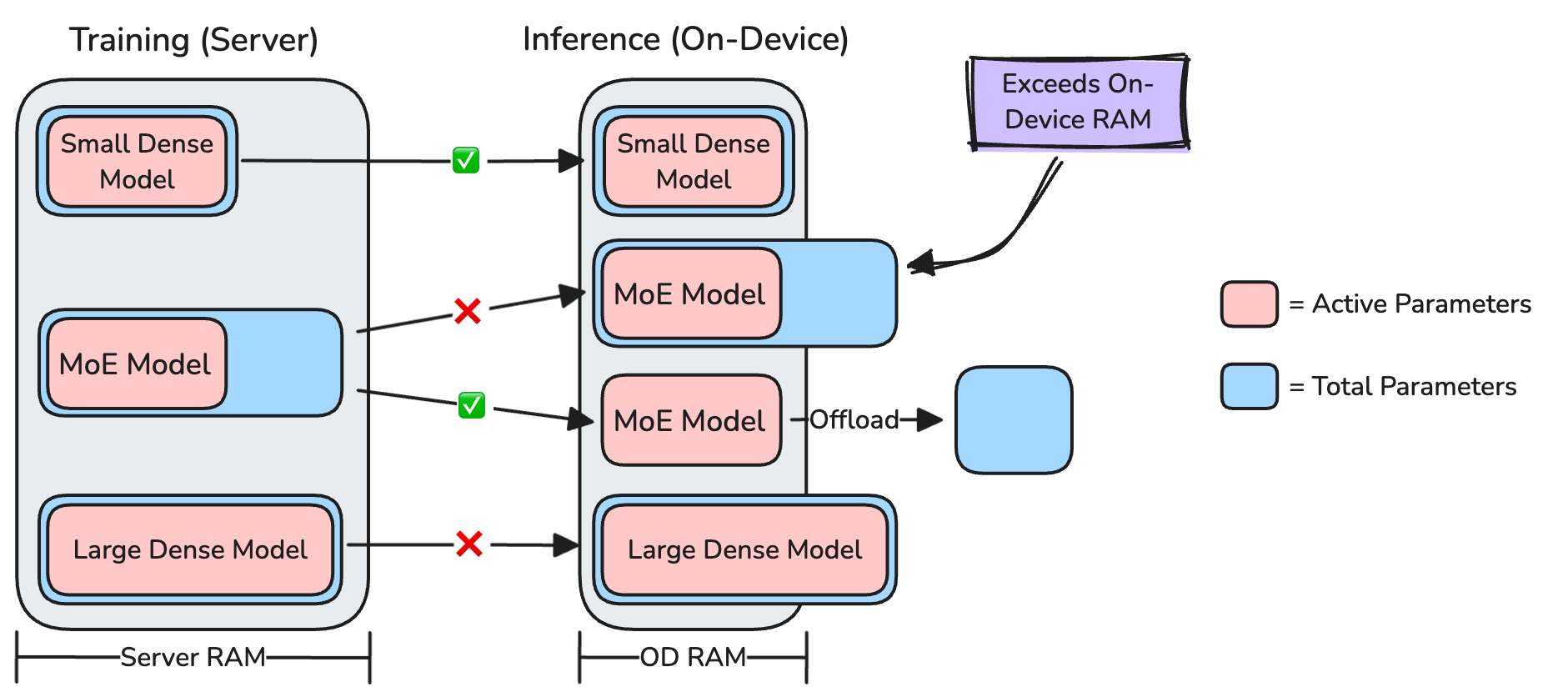}
    }
    \caption{Server-side training environment (left) compared to the memory-constraint inference environment (right), showing deployment restrictions for parameter heavy MoEs and large dense models on edge devices.}
    \label{fig:train-inference}
\end{figure}

\section{Introduction}
Mixture of Experts (short: MoEs) have been a popular extension of the transformer architecture \cite{vaswani2023attentionneed}, introducing the idea that each token of the input sequence is not fed through a single, dense network per layer, but a set of sub-networks, or ``experts''. To allow every input token to utilize a mixture of expert, the sub-networks are usually combined with a gating mechanism, which determines the contribution of each expert. 

In the general MoE setting, going back to \citet{6797059} and \citet{716791}, all experts are used to compute the final layer output. Building on top of this general architecture, sparse Mixture of Expert models have been proposed as a more compute-efficient alternative, only allowing a subset of experts to be activated for each token \cite{cai2024surveymixtureexperts}. 
Recently, many foundational models have adopted the MoE approach, such as Qwen \cite{bai2023qwentechnicalreport,yang2024qwen2technicalreport}, OLMoE \cite{muennighoff2024olmoeopenmixtureofexpertslanguage}, Mixtral \cite{jiang2024mixtralexperts}, Deepseek \citeyearpar{deepseekai2024deepseekv2strongeconomicalefficient}, \textit{inter alia}. 

In comparison to large-scale foundational Mixture of Expert models, optimized for highly parallelized server-side inference, in this work, we focus on small-scale foundational MoEs models deployed on edge devices\footnote{We focus on two model sizes: ``Wearables-sized'' models at 200M active parameters and ``Phone-sized'' models at 1.4B active parameters.}. As such, this comes with a set of challenges around single-sample, on-device inference, which can be classified into three categories: Quality, Memory and Latency.

\paragraph{Quality:} We tackle the fundamental research question if Mixture-of-Expert models can improve language modeling abilities over dense models at on-device scale. In comparison to previous work (e.g. \citet{jiang2024mixtralexperts}), we set up a truly fair comparison between MoEs and dense models. Here, we define a ``fair comparison'' of an MoE model against its dense counterpart by aligning for both, the same number of active parameters (i.e. FLOP aligned, short: \textit{FA}) and total parameters (i.e. parameter aligned, short: \textit{PA}). We further assume that a ``fair comparison'' between models should reduce confounding factors. Along those lines, we normalize models for training datasets, recipes, and architectures wherever possible. This way, we can make a clear performance attribution to the MoE component in isolation. In our evaluation, we show that MoE-style architectures improve the average language modeling performance by at least 2.35\% absolute across on-device model sizes. 
Based on these results, we propose a novel MoE model extension following the core intuition of ``expert specialization''. Using weight-decomposed experts, we show up to an additional 1.1\% language modeling improvements.

\paragraph{Memory/Latency:} For server-side models, language modeling ability presents the main dimension for model improvements. In the on-device context, however, we face two additional hard constraints: Memory and latency. As depicted in Figure \ref{fig:train-inference}, models trained in server environments, with loose memory and latency restrictions, face additional constraints for inference on edge-devices. While these restrictions are architecture independent, MoE-style models with a high total parameter count are more impacted. Luckily, the sparsity property of MoE architectures allows to circumvent this restriction by offloading unused experts, effectively reducing the model size in memory to the active parameter count (see Figure \ref{fig:train-inference}). Reducing the model memory through expert offloading, however, comes at the cost of 4-20x increased inference latency, since experts might need to be offloaded for every single token in the output sequence \cite{xue2024moeinfinityoffloadingefficientmoemodel}. To relax this memory/latency trade-off, we propose a novel ``block-wise expert selection'' loss, reducing expert offloads by 6x and, in turn, improving inference latency by 50\% compared to default offloaded MoEs.

\section{\methodname~Models}

\subsection{Sparse Mixture-of-Experts}
\label{sec:smoe}
\begin{figure}[t]
    \centering
    \resizebox{0.35\linewidth}{!}{
    \includegraphics{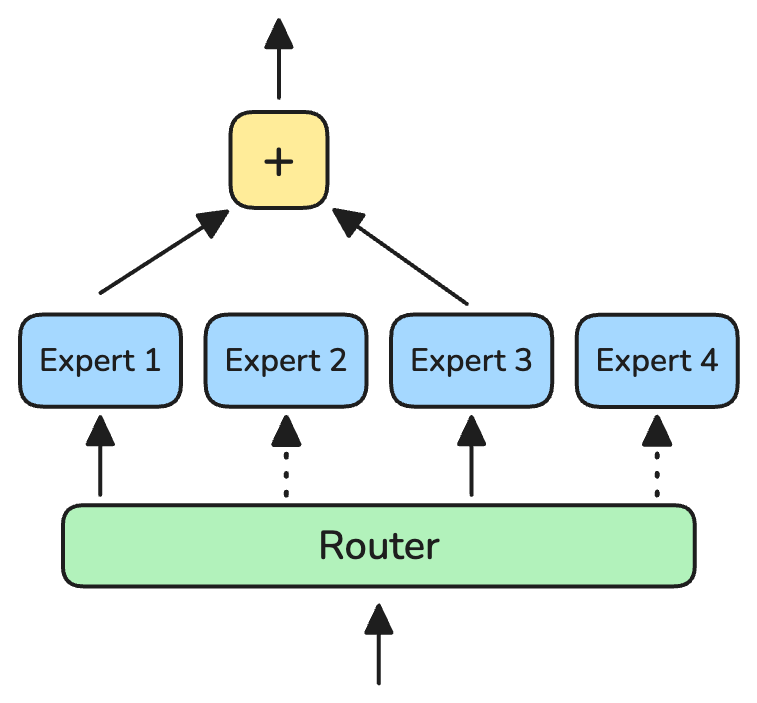}}
    \caption{Sparse Mixture-of-Experts architecture with Token Choice (TC) Routing and k=2}
    \label{fig:moe_overview}
\end{figure}

At the core of this work is the sparse Mixture-of-Expert (MoE) architecture, popularized by works such as GShard \cite{lepikhin2020gshardscalinggiantmodels} and Switch Transformers \cite{fedus2022switchtransformersscalingtrillion}. While MoEs can generally be implemented for different parts of the architecture, the most common approach is to replace the single dense feed-forward layer with a router component and multiple experts (see Figure \ref{fig:moe_overview}). Selecting a discrete subset of experts at each step, sparse MoE models can be defined by their active parameters (FLOPs) and total parameters (model size in memory). The resulting FLOP-to-parameter ratio directly translates to increased training and inference efficiency, without sacrificing model performance. To find a suitable subset of experts, different expert routing paradigms have been established, either selecting experts per token (token choice or ``TC'') \cite{shazeer2017outrageouslylargeneuralnetworks} or per expert (expert choice or ``EC'') \cite{zhou2022mixtureofexpertsexpertchoicerouting}. Here, we use the token choice expert routing paradigm (illustrated in Figure \ref{fig:moe_overview}) following the findings in OLMoE \cite{muennighoff2024olmoeopenmixtureofexpertslanguage}, showing that EC does not bring clear improvements for text-only models. \\
Please note that from here on out, we will refer to sparse MoEs as solely ``MoEs'' for brevity. However, all evaluated models in this paper are sparse versions of Mixture-of-Expert models.

\subsection{Weight-Decomposed Experts}
\label{sec:lora}
\begin{figure}
    \centering
    \resizebox{0.65\linewidth}{!}{
    \includegraphics{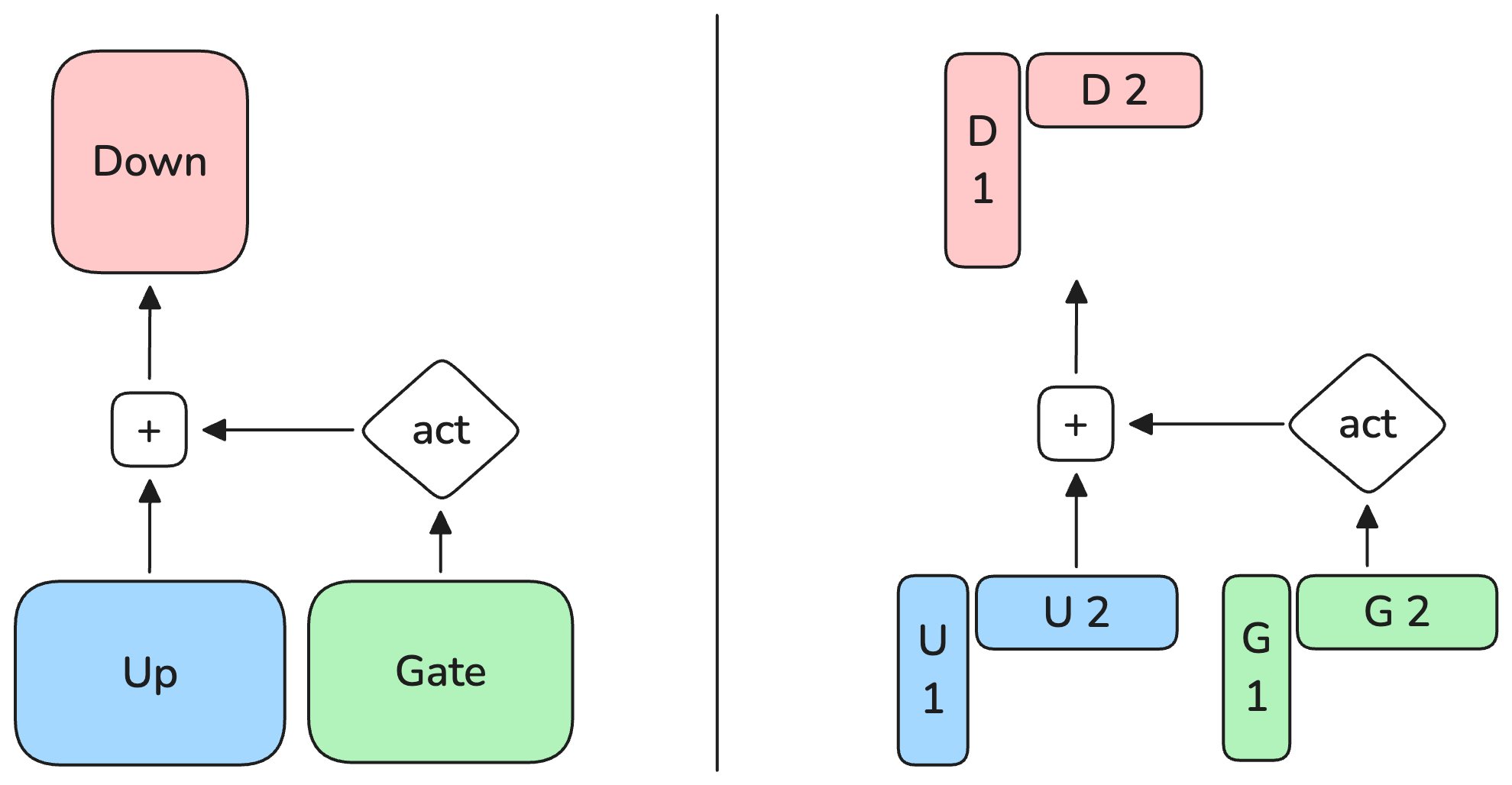}}
    \caption{Feed Forward Layer: Standard (left) and Weight-Decomposed (right).}
    \label{fig:lora-moe}
\end{figure}

To reduce the naturally large total parameter count of MoE-style models, we propose a lightweight definition of experts using matrix weight decompositions (``WD'') similar in spirit to Low Ranking (``LoRa'') adapters \cite{hu2021loralowrankadaptationlarge}. Intuitively, individual experts are intended to ``specialize'' towards a subset of, ideally, $\frac{1}{\# Experts}$ tokens. Based on this intuition, we replace expert matrices of shape $n \times m$ with weight decompositions of shape $n \times r$ and $r \times m$ as shown in Figure \ref{fig:lora-moe} and defined in Equation \ref{eq:wd}:
\begin{equation}
\label{eq:wd}
    M_{n \times m} \approx L_{n \times r} \times R_{r \times m}
\end{equation}

Here, the original matrix $M$ is replaced by $L$ and $R$, with $r \ll n$ and $r \ll m$. In preliminary experiments, we test multiple reduction factors for $r$ and find that a decomposition of half the hidden dimension results in the best trade-off between parameter reduction and model performance. Weight decomposed models are from here on out prefix with a \textit{WD} term. To ensure a paramter-aligned comparison, we adjust the number of heads and layers as further elaborated on in section \ref{sec:model_config}. 

\subsection{Block-wise Expert Selection}
\label{sec:BlES}
We now explore the second restrictive dimension of MoEs for on-device use cases: Memory and Latency. Multiple lines of research have previously explored inference-time optimizations using predictive expert offloading and bitwidth adaptations, such as EdgeMoE \cite{yi2023edgemoefastondeviceinference}, Mixtral \cite{eliseev2023fastinferencemixtureofexpertslanguage} and DeepSpeed \cite{aminabadi2022deepspeedinferenceenablingefficient}. Here we explore the expert offloading problem from a new vantage point, proposing a ``Block-wise Expert Selection'' (BlES) training loss term to reduce the number of expert replacements. Our BlES loss is thereby closely related to the expert load balancing loss proposed in \citet{fedus2022switchtransformersscalingtrillion}:

Let $R$ be a router logits tensor with shape $(B, T, E)$. With $B$ as the batch-dimension, $T$ as the sequence length and $E$ as the expert dimension. We compute the routing weights $W$ by applying the softmax function to $R$, scaled by a temperature parameter $\tau$ as:
\begin{equation}
    W = \text{softmax}(\tau R)
\end{equation}

In the non-differentiable part of the loss, we select the top-k experts $K$ for each token based on the routing weights $W$. Let $S$ be the selected experts tensor with shape $(B, T, K)$ following
\begin{equation}
S = top\_k(W, K)
\end{equation}

We then compute the number of hard expert replacements $H$ by comparing consecutive tokens' expert assignments as:
\begin{equation}
\label{eq:H_1}
\begin{aligned}
H_e &= \sum_{b=1}^{B}\sum_{t=1}^{T-1} |(S_{[b, t+1]} == e) - (S_{[b, t]} == e)| \\
H &= \sum_{e=1}^{E} H_e
\end{aligned}
\end{equation}

where $e$ is the expert index and $S_{[b, t]} == e$ is $1$ if expert $e$ is one of the top-k candidates for token $t$. This approach counts every expert replacement twice ($1\rightarrow0$ for the active expert and $0\rightarrow1$ for newly active expert). As a result, we divide $H$ by two and normalize by the batch-size, top-k and number of tokens as follows:
\begin{equation}  
\label{eq:H_2}
H_{norm} = \frac{\left \lfloor \frac{H}{2} \right \rfloor}{B \cdot K \cdot (T - 1)}
\end{equation}

To keep the overall loss term differentiable, we compute a soft expert selection $L$ by combining the per-expert probability differences between consecutive tokens along the token dimension $T$.
With $L_{norm}$ as the normalized soft expert selection, we compute:
\begin{equation}  
\begin{aligned}
L &= \sum_{b=1}^{B} \sum_{t=1}^{T-1} \sum_{e=1}^{E} |W_{b,t+1,e} - W_{b,t,e}| \\
L_{norm} &= \frac{L}{B \cdot T}
\end{aligned}
\end{equation}
The final loss is defined as product of the hard and the soft expert selection loss.
\begin{equation}
loss = H_{norm} \cdot L_{norm}
\end{equation}

\begin{figure}[h!]
    \centering
    \resizebox{0.75\linewidth}{!}{
    \includegraphics{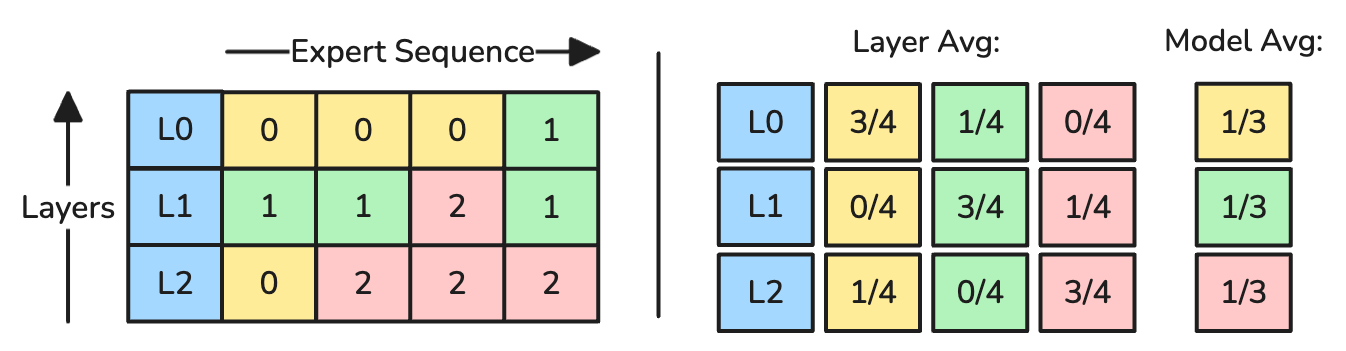}}
    \caption{Example expert selection (for simplicity, k=1) for individual layers and the complete model.}
    \label{fig:layer_wise_vs_model_wise_expert_balance}
\end{figure}

As described above, the block-wise expert selection loss is defined on sequence level. We adjust the standard load balancing loss \cite{fedus2022switchtransformersscalingtrillion} to also operate on sequence level (following \citet{lin2024momaefficientearlyfusionpretraining}) to avoid loss inconsistencies, allowing the model to ``cheat''.  For example, using 2 experts and 2 layers, the loss function can be exploited by consistently selecting expert 0 in layer 0 and expert 1 in layer 1, hence having a perfect 50:50 load balancing loss at the model level, as well as a minimal BlES on sequence level. See Figure \ref{fig:layer_wise_vs_model_wise_expert_balance} for a visualization of this example using 3 layers and 3 experts.

\section{Evaluation}
\label{sec:eval}

\subsection{Model Setup and Training Recipes}
\label{sec:model_config}
We compare two on-device sizes: ``Phone-sized'' ($\sim$1-3B parameters) and ``Wearable-sized'' ($\sim$100-300M parameters) as well as three architectures: Dense, MoE and WD MoE, all presented in Table \ref{tab:model_configs}. We further train the standard MoE architecture with our novel ``Block-wise Expert Selection'' (BlES) loss\footnote{The BlES model uses the standard MoE model architecture and is not separately mentioned in Table \ref{tab:model_configs}.}.
All models are based on the Llama3 architecture, with the additional MoE component consisting of eight total experts, with two active for every token. We follow standard approaches provided in the Huggingface codebase for the expert implementation \cite{wolf2020huggingfacestransformersstateoftheartnatural}. We keep model all hyper-parameters as constant as possible while aligning dense and MoE models along the active and total parameter counts. When in doubt, we follow the findings in \citet{liu2024mobilellmoptimizingsubbillionparameter} and select depth over breadth.

\begin{table}[h]
    \centering
    \begin{tabular}{l|r|rrrrrr}
        Model & Params & L & H & Hid & Seq & Steps & Bsz \\
        \hline
        \multicolumn{8}{c}{\textbf{Phone-sized models, $\sim$1B-3B Parameters}} \\
        \hline
        Dense & 1.50B & 16 & 32 & 2048 & 2048 & 310k & 2048 \\
        MoE & 1.37B (3.75B) & 24 & 18 & 1440 & 2048 & 310k & 2048 \\
        ~~+ WD & 1.42B (3.65B) & 26 & 20 & 1600 & 2048 & 310k & 2048 \\
        Dense & 3.61B & 28 & 24 & 3072 & 2048 & 310k & 2048 \\
        \hline        
        \multicolumn{8}{c}{\textbf{Wearable-sized models, $\sim$100-200M Parameters}} \\
        \hline
        Dense & 189M & 19 & 8 & 512 & 2048 & 310k & 2048 \\
        MoE & 188M (377M) & 19 & 8 & 432 & 2048 & 310k & 2048 \\
        ~~+ WD & 188M (377M) & 32 & 10 & 400 & 2048 & 310k & 2048 \\
        Dense & 380M & 29 & 12 & 768 & 2048 & 310k & 2048 \\
    \end{tabular}
    \caption{On-device model candidates. Params = \#Active (\#Total) Parameters, L = Layers, H = Self-Attention Heads, Hid = Hidden size, Seq = Sequence length, Bsz = effective batch-size}
    \label{tab:model_configs}
\end{table}

\subsection{Training Datasets}
\label{sec:datasets}
To pre-train all models using the FineWeb Education dataset (FW-edu, \citet{penedo2024finewebdatasetsdecantingweb}), a 1.4 trillion token text dataset provided by Huggingface \cite{wolf2020huggingfacestransformersstateoftheartnatural}. Compared to other popular, open-source pre-training datasets, such as RedPajamas \cite{together2023redpajama, weber2024redpajama}, FW-edu represents a smaller scale, yet high-quality, general purpose language dataset, filtered with Llama-70B educational prompts.

\subsection{Metrics and Benchmarks}
\label{sec:metrics}
To evaluate the \textbf{language modeling performance} we use the public EleutherAI LM eval harness and nine language modeling evaluations \cite{eval-harness}, namely, MMLU, AGI-English, Arc-challenge and -easy, BoolQ, PIQA, SIQA, HellaSwag and WinoGrande. We pick this subset in accordance with Llama3 \cite{grattafiori2024llama3herdmodels} and MobileLLM \cite{liu2024mobilellmoptimizingsubbillionparameter} evaluations. We exclude long-context evaluations (e.g. SQuAD, DROP, QuAC), due to our sequence length restriction of 2048. To evaluate our \textbf{Block-wise Expert Selection (BlES)} loss, we show two offloading-specific metrics: The Expert Replacement Ratio (ExRep) and optimal expert balance. Specifically, the Expert Replacement is defined along the lines of equations \ref{eq:H_1} and \ref{eq:H_2}, calculating the percentage of realized replacements. Regarding the optimal expert balance, we calculate the average per-layer delta between the uniform distribution and the realized expert balance. Lastly, to investigate the model candidates' \textbf{memory and latency performance}, we show the per-token model latency (i.e. the realized generation speed) and peak memory.

\begin{table}[t]
  \centering
  \resizebox{1.0\linewidth}{!}{
  \begin{tabular}{l|r|rrrrrrrrrr|r}
    Model & Params & MMLU & AGI-E & Arc-C & Arc-E & BoolQ & PIQA & SIQA & HellaS & OBQA & WinoG & Avg \\
    \hline
    \multicolumn{13}{c}{\textbf{Random Baseline}} \\
    \hline
    Random & -- & 24.53  & 16.07  & 21.08  & 25.25  & 51.07  & 51.74  & 33.11  & 26.31  & 29.40  & 50.83  & 32.94 \\
    \hline
    \multicolumn{13}{c}{\textbf{Phone-sized models, $\sim$1B-3B Parameters}} \\
    \hline
    Dense & 1.50B & 24.78 & 17.99 & 36.95 & 74.03 & 59.08 & 74.54 & 41.76 & 59.88 & 41.20 & 57.54 & 48.78 \\
    MoE & 1.37B (3.75B) & \underline{25.96}  & 17.65  & 42.58  & 76.77  & 60.89  & 75.52  & 42.12  & 65.07  & 42.40  & \underline{62.35}  & 51.13 \\
    ~~+ BlES & 1.37B (3.75B) & 25.40 & 17.50 & 41.55 & \underline{77.02} & 62.81 & 76.06 & 41.91 & 63.14 & 42.60 & 59.04 & 50.70 \\
    ~~+ WD & 1.42B (3.65B) & 23.90  & \underline{\textbf{18.20}}  & \underline{43.69}  & 76.81  & \underline{\textbf{66.76}}  & \underline{76.39}  & \underline{\textbf{45.14}}  & \underline{66.51}  & \underline{42.80}  & 62.04  & \underline{52.22} \\
    Dense  & 3.61B & \textbf{26.41} & 16.82 & \textbf{44.54} & \textbf{77.9} & 65.87 & \textbf{77.48} & 43.3 & \textbf{67.18} & \textbf{45.00} & \textbf{63.46} & \textbf{52.80} \\
    \hline
    \multicolumn{13}{c}{\textbf{Wearable-sized models, $\sim$100-200M Parameters}} \\
    \hline
    Dense & 189M & 22.9 & 16.82 & 23.29 & 56.82 & 57.09 & 64.15 & 37.82 & 36.36 & 32.8 & 50.99 & 39.90 \\
    MoE & 188M (377M) & \underline{\textbf{25.27}} & 17.37 & 27.9 & \underline{63.09} & 58.39 & 69.04 & 39.61 & 44.09 & \underline{\textbf{34.4}} & 53.03 & 43.22 \\
    ~~+ BlES & 188M (377M) & 24.27 & \underline{17.58} & 24.83 & 58.84 & \underline{\textbf{59.82}} & 66.49 & 38.64 & 39.70 & 33.40 & 49.96 & 41.35 \\
    ~~+ WD & 188M (377M) & 23.64 & 17.16 & \underline{28.58} & 62.58 & 57.13 & \underline{\textbf{69.31}} & \underline{\textbf{40.28}} & \underline{46.15} & 33.20 & \underline{\textbf{54.38}} & \underline{\textbf{43.24}} \\
    Dense & 380M & 24.79 & \textbf{17.86} & \textbf{28.92} & \textbf{64.35} & 52.02 & 69.21 & 39.97 & \textbf{46.53} & 33.80 & 51.62 & 42.91 \\
    \hline
    \multicolumn{13}{c}{\textbf{Public Baselines across Model Sizes}} \\
    \hline
    MobLLM \citeyearpar{liu2024mobilellmoptimizingsubbillionparameter}& 135M & 23.02 & 17.45 & 19.97 & 46.38 & 60.34 & 64.96 & 38.08 & 38.17 & 28.40 & 52.57 & 38.93 \\
    MobLLM \citeyearpar{liu2024mobilellmoptimizingsubbillionparameter}& 350M & 26.33 & 17.47 & 23.89 & 56.4 & 61.96 & 68.88 & 39.87 & 49.57 & 31.00 & 57.38 & 43.28 \\
    \hline
    Llama3.2 \citeyearpar{grattafiori2024llama3herdmodels} & 1.4B & 36.92  & 18.80  & 31.31  & 65.40  & 63.61  & 74.54  & 42.84  & 47.74  & 26.20  & 60.06  & 46.70  \\
    Llama3.2 \citeyearpar{grattafiori2024llama3herdmodels} & 3.6B & 54.01  & 22.53  & 42.32  & 74.41  & 72.81  & 76.71  & 47.13  & 55.32  & 31.20  & 69.30  & 54.50 \\
    OLMoE \citeyearpar{muennighoff2024olmoeopenmixtureofexpertslanguage} & 1.68B (6.92B) & 25.74  & 17.19  & 40.87  & 74.20  & 60.52  & 74.70  & 44.37  & 60.38  & 38.40  & 58.72  & 49.50 \\
  \end{tabular}}
  \caption{Model comparison on zero shot LM evaluations. Params = \#Active (\#Total) Parameters, BlES = Block-wise Expert Selection, WD = Weight-Decomposed, MobLLM = MobileLLM. Public baselines are evaluated using the EleutherAI LM eval harness \citeyearpar{eval-harness}.
  }
  \label{tab:comparison_1b}
\end{table}

\subsection{Results}

\subsubsection{Language Modeling Performance}
\label{sec:lm}
Our results on the language modeling task are presented in Table \ref{tab:comparison_1b}. We show a random baseline in the top row of the table, followed by the main model comparisons. The MoE-based results are framed by two rows of dense model candidates. On top of each sub-table, we show the FLOP aligned model (short: \textit{FA}), at the bottom we present the parameter aligned (short: \textit{PA}) dense model. For the MoE candidates, we show the standard MoE followed by the BlES and weight decomposed (WD) versions. In the bottom sub-table we show additional models from the literature to put our results into context\footnote{Previously published models are also evaluated using the EleutherAI LM eval harness, but not aligned for confounding factors and, hence, not directly comparable.}.

\paragraph{Phone-sized models:} We show that all MoE model candidates outperform the random baseline by a large margin and consistently improve over the FA dense model by at least 2\%. Comparing individual tasks, we find that for MMLU and AGI-English, all tested models only provide minor gains compared to the random baseline, showing clear potential for further improvements in this area. Regarding all other evaluation tasks, clear improvements are observed. Between MoE models, the weight decomposed model performs best overall, while for individual metrics the top-performing candidate varies. We also find a minor performance regression when using the block-wise expert selection loss. Compared to the PA dense model, MoE candidates perform better in 3 out of 10 metrics, falling only about half a percent short on average. Putting our observed model performances into the context of previously published models (1B and 3B Llama3.2, OLMoE 1B-7B), we find that the MoE model candidates outperform the FA Llama 3.2 1B and OLMoE models, however, can not reach the PA Llama 3.2 3B performance. We believe that this clearly shows that our MoE-style models are competitive to top open source candidates.

\paragraph{Wearable-sized models:} The wearbale-sized evaluation shows generally similar trends. All MoE candidates outperform the random baseline and FA dense model. MMLU and AGI-English results are insignificantly above the random baseline, while all other tasks show meaningful improvements. The weight-decomposed model achieves the best MoE performance, this time even outperforming the PA dense model. At wearable-scale, at least one of the MoE models outperforms the PA dense model in 6 of 10 tasks. Looking at the comparison to the previously published MobileLLM model, we see improvements at the 125M and 350M parameter scale. Again, the BlES model shows a slight performance drop compared to the standard MoE setup.

\begin{table}[h!]
    \centering
    \begin{tabular}{l|rrr}
        Model& ExRep ($\downarrow$) & Tok/s Gen ($\uparrow$) & $\Delta \text{Uni}$ ($\downarrow$)  \\
        \hline
        MoE & 43.82 & 15.02 & \textbf{9.60} \\
        ~~+ BlES & \textbf{6.55} & \textbf{23.10} & 9.67 \\
    \end{tabular}
    \caption{Impact of the BlES Loss on Expert Replacement (in percent), generation speed (token/second), and diversion from the uniform expert distribution (in percent) $\downarrow$ =  lower is better, $\uparrow$ = higher is better.}
    \label{tab:BlES}
\end{table}

\subsubsection{Offload Efficiency}
\label{sec:BlES_results}
As previously shown in Figure \ref{fig:train-inference}, executing MoE models on-device requires offloading experts to stay within memory constraints. This necessity, however, causes significant latency regressions, rooted in the added offloading overhead. Let $E$ be the set of experts, $S$ the set of selected experts, and $N = E \setminus S$ the set of non-selected experts. For each token in the output sequence, the following offloading logic is applied to ensure the number of experts in GPU memory never exceeds the number of active experts:
\begin{equation}
\begin{aligned}
& \text{If } S \neq S_{prev}: \\
& \quad \forall e \in N \to \text{CPU} \\
& \quad \forall e \in S \to \text{GPU} \\
& \quad S_{prev} \leftarrow S
\end{aligned}
\label{eq:offload}
\end{equation}

Since the expert selection and, hence, offloading frequency is data-dependent, we use a $100$ sample subset of the C4 dataset \cite{JMLR:v21:20-074} as a proxy for general text data. Table \ref{tab:BlES} presents the results of this evaluation along three dimensions: The expert replacement percentage (ExRep), the realized inference speed in tokens per second (the full set of on-device benchmarks, putting the generation speeds into context, is presented in section \ref{sec:od}), and the model diversion from the ideal uniform expert balance ($\Delta \text{Uniform}$). Comparing the standard MoE model with our BlES extension, we find that the additional loss term causes a significant reduction in expert replacements, reducing the number of expert switches by over 6 times. This also directly converts into a real-world generation speed improvement of over 1.5x. Looking at the third metrics in Table \ref{tab:BlES}, we observe a minor increase in the optimal expert balancing metric of less than 1\% relative\footnote{Please note that the shown inference latency improvement is batch-size dependent.}. 

Besides the quantitative results in Table \ref{tab:BlES}, we show a qualitative example in Figure \ref{fig:qual_err}. Compared to the standard MoE model (bottom), the BlES loss extended model (top) effectively reduces the number of expert replacements from 21$\rightarrow$11, while conserving expert diversity (both models actively use 6 out of the 8 experts). 

\begin{figure}[h!]
    \centering
    \begin{tikzpicture}
    \begin{axis}[
        ybar,
        enlargelimits=0.1,
        bar width = 9pt,
        legend style={at={(0.5,-0.4)},
        anchor=north,legend columns=-1},
        ylabel={$\Delta$Uniform ($\downarrow$)},
        height=3.8cm,
        width=0.7\linewidth,
        xticklabel style={rotate=0},
        symbolic x coords={1-3, 4-6, 7-9, 10-12, 13-15, 16-18, 19-21, 22-24},
        xtick=data,
    ]
    \addplot[replace_alt, fill=replace_alt] coordinates {(1-3, 6.98)	(4-6, 5.92)	(7-9, 11.37)	(10-12, 10.5)	(13-15, 10.98)	(16-18, 10.4)	(19-21, 8.76)	(22-24, 11.91)};
    \addplot[replace, fill=replace] coordinates {(1-3, 15.06)	(4-6, 13.19)	(7-9, 7.61)	(10-12, 8.51)	(13-15, 8.31)	(16-18, 8.84)	(19-21, 7.39)	(22-24, 8.43)};
    \legend{MoE, MoE + BlES}
    \end{axis}
    \end{tikzpicture}
    \caption{Per layer analysis of the divergence of the expert routing from the uniform expert distribution. Large values indicate expert collapse and use of a pseudo-dense layer.}
    \label{fig:layer_delta}
\end{figure}
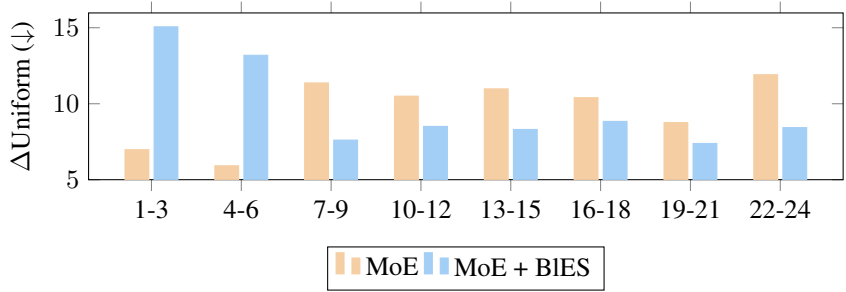

Furthermore, to get a better understanding of the per-layer impact of the BlES loss, we plot the layer-wise expert balance analysis in Figure \ref{fig:layer_delta}. We find that when using the blocked expert selection, a larger expert divergence is observed in lower layers, while the standard MoE model shows a generally higher expert balance divergence in higher layers. While we don't have a clear understanding of the reasoning and impact of these differences, we believe that higher expert diversity in later layers seems preferable, given the general intuition that lower layers encode more local, syntactic information, while higher layers represent more global and semantic structures.

\begin{figure}
    \resizebox{\linewidth}{!}{
        \begin{tikzpicture}[scale=0.75]
            \foreach \y in {1,...,35}{
                \node[minimum size=6mm] at (\y, 0) {\y};
                }
        
          \foreach \y [count=\n] in {
            {1,0,0,0,0,0,0,0,0,1,1,1,0,0,0,1,1,1,1,1,0,0,0,1,0,0,0,0,0,0,1,1,0,0,0},
            {0,0,0,0,0,0,0,0,0,0,0,0,0,0,0,0,0,0,0,0,0,0,0,0,0,0,0,0,0,0,0,0,0,0,0},
            {0,0,0,0,0,0,0,0,0,0,0,0,0,0,0,0,0,0,0,0,0,0,0,0,0,0,0,0,0,0,0,0,0,0,0},
            {0,0,0,1,1,1,1,1,1,0,0,0,1,1,1,0,0,0,0,0,0,0,0,0,0,0,0,0,0,0,0,0,0,0,0},
            {1,1,1,1,1,1,1,1,1,1,1,1,1,1,1,1,1,1,1,1,1,1,1,1,1,1,1,1,1,1,1,1,1,1,1},
            {0,0,0,0,0,0,0,0,0,0,0,0,0,0,0,0,0,0,0,0,1,1,0,0,0,0,0,0,0,0,0,0,0,0,0},
            {0,0,0,0,0,0,0,0,0,0,0,0,0,0,0,0,0,0,0,0,0,0,0,0,1,1,1,1,1,1,0,0,1,1,1},
            {0,1,1,0,0,0,0,0,0,0,0,0,0,0,0,0,0,0,0,0,0,0,1,0,0,0,0,0,0,0,0,0,0,0,0},
            } {
              \foreach \x [count=\m] in \y {
               \pgfmathsetmacro\result{\x*100}
                \node[fill=replace!\result!no_replace, minimum size=6mm, text=black] at (\m,-\n) {\x};
              }
            }
        
          \foreach \a [count=\i] in {$E_1$, $E_2$, $E_3$, $E_4$, $E_5$, $E_6$, $E_7$, $E_8$} {
            \node[minimum size=6mm] at (0,-\i) {\a};
          }
        \end{tikzpicture}
        }

    \vspace{0.5cm}
        
    \resizebox{\linewidth}{!}{
        \begin{tikzpicture}[scale=0.75]
        
        
          \foreach \y [count=\n] in {
            {0,1,1,1,1,1,1,1,1,0,1,0,0,0,1,1,1,1,1,0,1,1,1,1,1,1,1,1,1,1,1,1,1,1,1},
            {1,0,1,1,1,1,1,1,0,0,0,1,1,1,1,1,1,1,0,1,0,0,0,1,1,1,0,1,1,1,0,1,1,1,1},
            {0,0,0,0,0,0,0,0,0,0,0,0,0,0,0,0,0,0,0,0,0,0,0,0,0,0,1,0,0,0,0,0,0,0,0},
            {1,0,0,0,0,0,0,0,0,1,0,1,1,0,0,0,0,0,1,1,0,0,0,0,0,0,0,0,0,0,0,0,0,0,0},
            {0,1,0,0,0,0,0,0,0,0,1,0,0,0,0,0,0,0,0,0,0,1,1,0,0,0,0,0,0,0,1,0,0,0,0},
            {0,0,0,0,0,0,0,0,0,0,0,0,0,0,0,0,0,0,0,0,0,0,0,0,0,0,0,0,0,0,0,0,0,0,0},
            {0,0,0,0,0,0,0,0,1,1,0,0,0,1,0,0,0,0,0,0,1,0,0,0,0,0,0,0,0,0,0,0,0,0,0},
            {0,0,0,0,0,0,0,0,0,0,0,0,0,0,0,0,0,0,0,0,0,0,0,0,0,0,0,0,0,0,0,0,0,0,0},
            } {
              \foreach \x [count=\m] in \y {
               \pgfmathsetmacro\result{\x*100}
                \node[fill=replace_alt!\result!no_replace_alt, minimum size=6mm, text=black] at (\m,-\n) {\x};
              }
            }
        
          \foreach \a [count=\i] in {$E_1$, $E_2$, $E_3$, $E_4$, $E_5$, $E_6$, $E_7$, $E_8$} {
            \node[minimum size=6mm] at (0,-\i) {\a};
          }
        \end{tikzpicture}
    }
    \caption{Example expert replacements. 1 = Active Expert, 0 = Inactive Expert. Top: BlES, Bottom: MoE.}
    \label{fig:qual_err}
\end{figure}

\subsubsection{On-Device Benchmarks}
\label{sec:od}
We now evaluate the model candidates along the two main on-device dimensions, namely, latency and peak memory.
Given that on-device models are oftentimes executed in either CPU based environments or using proprietary accelerators, we compare model latency in both, CPU and GPU environments\footnote{Please note that our evaluations are executed in a server environment and actual on-device accelerator numbers might vary.}. Furthermore, despite a variety of inference-optimizations available across different modeling frameworks and code bases (e.g. EdgeMoE \cite{yi2023edgemoefastondeviceinference}), this paper targets training-time improvements. As a result, we use standard inference code provided in the Huggingface Transformers library \cite{wolf2020huggingfacestransformersstateoftheartnatural} and the gpt-fast codebase \cite{gpt-fast} without further inference optimizations.

Table \ref{tab:od_measures} shows our results along four dimensions: (1) The language modeling performance, as previously shown in Table \ref{tab:comparison_1b}, (2) the model inference speed in tokens per second, measured across 128 tokens in CPU and GPU environments, (3) the model peak memory after 128 token generations in GB of RAM and (4) the suitability of the model for on-device inference (in line with Figure~\ref{fig:train-inference}). 

Besides the previously shown model candidates, we add an additional standard MoE offloading setup following equation \ref{eq:offload}, indicated as ``Offl'', besides the ``BlES'' offloaded model.

\paragraph{Latency:} Looking at the generation latency, we find that on CPU, the FA dense model achieves the highest token per second generation, MoE model candidates are slightly slower, and the PA dense model regresses the generation speed by 2x. On GPU, MoE models generally produce less tokens per second than dense models, mainly caused by the deeper architecture (see layer comparisons in Table \ref{tab:model_configs}). Looking at offloading enabled models, further slowdowns can be observed due to expert offloading delays. Comparing the standard offloaded MoE model against our BlES offloaded model, we find the 1.5x speed-up in generation speed, as previously presented in Table \ref{tab:BlES}. 

To put these results into context, inference-based offloading strategies, such as  \citet{eliseev2023fastinferencemixtureofexpertslanguage} and \citet{aminabadi2022deepspeedinferenceenablingefficient} achieve a 2-3x and 5.5x generation latency reduction at the most comparable model size, which is still significantly larger than our on-device sized models. Furthermore, while orthogonal to our train-time improvements, inference-time offloading methods can oftentimes not be used in on-device centered scenarios, due to their additional modeling components required to predict future expert use.

\paragraph{Peak Memory:} We find that without expert offloading, the generation peak memory of the MoE model candidates is, as expected, comparable to the PA dense model. Using expert offloading, peak memory during generation is reduced to the FA dense model, given that only active parameters are kept in memory, making only offloaded MoE models true on-device candidates (see \twemoji{check mark button} in the right-most column).

\begin{table}[t]
    \centering
    \begin{tabular}{l|r|rr|r|c}
        Model & \multicolumn{1}{c|}{LM Eval} & \multicolumn{2}{c|}{Latency} & \multicolumn{1}{c|}{Mem} & \\
        \hline
        Setup & Avg & \multicolumn{2}{c|}{Gen (tok/sec)} & Gen & \twemoji{mobile phone}\\
        \hline
         Metric & \% & CPU & GPU & GB & \twemoji{check mark button} / \twemoji{cross mark} \\
        \hline
        Dense & 48.78 & 4.47 & 73.10 & 5.8 & \twemoji{check mark button} \\
        MoE & 51.13 & 4.30 & 40.60 & 14.7 & \twemoji{cross mark}  \\
        ~+ WD & 52.22  & 3.85 & 33.50 & 14.2 & \twemoji{cross mark}  \\
        ~+ Offl & 51.13  & 4.30 & 15.02 &5.4 & \twemoji{check mark button} \\
        ~+ BlES & 50.70 & 4.30 & 23.10 & 5.4 & \twemoji{check mark button} \\ 
        Dense & 52.80 & 1.77 & 42.60 & 14.0 & \twemoji{cross mark}  \\
    \end{tabular}
    \caption{On-device benchmarks. Gen = Generation of 128 tokens (1 token prefill), Offl = Offloaded, BlES=Block-wise Expert Selection. Mem = Peak GPU memory. \twemoji{mobile phone}= Phone-sized, assuming <6GB of RAM use (e.g. iPhone 12 Pro).}
    \label{tab:od_measures}
\end{table}

\subsubsection{On-Device Expert Ablations}
\begin{figure}[t]
    \centering
    \begin{tikzpicture}
        \begin{groupplot}[
            group style={group size=3 by 2, horizontal sep=1.2cm, vertical sep=1.0cm},
            xtick=data,
            height=0.22\linewidth,
            width=0.25\linewidth
            ]
            \nextgroupplot[font=\small, ymin=46.5, ymax=52.5, ylabel style={align=center}, ylabel={\textbf{Active Experts}}, xticklabels={1, 2, 4, 8}, xlabel style={align=center}]
            \addplot coordinates {
                (0, 47.44)
                (1, 49.75)
                (2, 51.23)
                (3, 51.84)
            };
            \nextgroupplot[font=\small, ymin=10, ymax=40, xticklabels={1, 2, 4, 8}]
            \addplot[color=red, mark=*] coordinates {
                (0, 37.64)
                (1, 28.13)
                (2, 20.84)
                (3, 13.38)
            };
            \nextgroupplot[font=\small, ymin=3, ymax=30, xticklabels={1, 2, 4, 8}, xlabel style={align=center}]
            \addplot[color=brown, mark=*] coordinates {
                (0, 14.7)
                (1, 14.7)
                (2, 14.7)
                (3, 14.7)
            };
            
            \nextgroupplot[font=\small, ymin=46.5, ymax=52.5, ylabel style={align=center}, ylabel={\textbf{Total Experts}}, xticklabels={2, 4, 8, 16}, xlabel={Avg LM evals}]
            \addplot coordinates {
                (0, 48.7)
                (1, 49.48)
                (2, 49.75)
                (3, 50.64)
            };

            \nextgroupplot[font=\small, ymin=10, ymax=40, xticklabels={2, 4, 8, 16}, xlabel style={align=center}, xlabel={Gen Speed (tok/s)}]
            \addplot[color=red, mark=*] coordinates {
                (0, 29.62)
                (1, 29.77)
                (2, 29.13)
                (3, 29.6)
            };
            
            \nextgroupplot[font=\small, ymin=3, ymax=30, xticklabels={2, 4, 8, 16}, xlabel style={align=center}, xlabel={Peak Mem (GB)}]
            \addplot[color=brown, mark=*] coordinates {
                (0, 5.3)
                (1, 8.5)
                (2, 14.7)
                (3, 27.5)
            };
        
            \end{groupplot}
    \end{tikzpicture}
    \caption{Active (top) and total (bottom) expert ablations of the 1.4B MoE model after 50,000 steps ($\sim$210B tokens)}
    \label{fig:exp_ablation}
\end{figure}
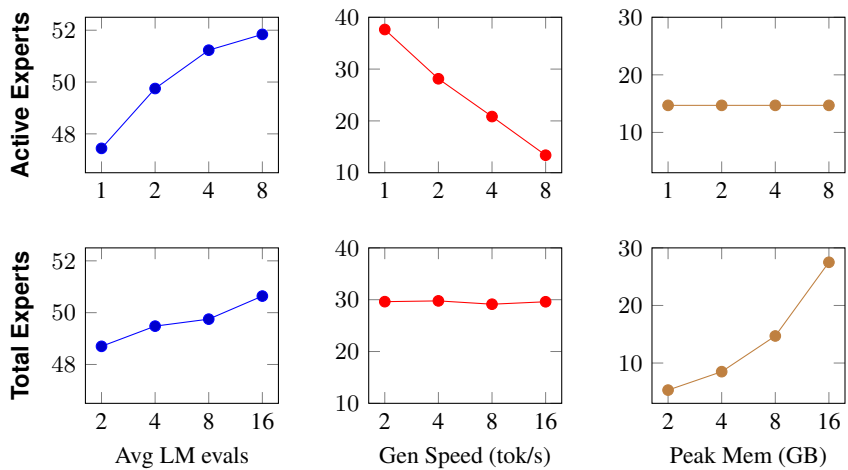

In the previous sections, we followed the standard MoE setup with two active and eight total experts. Going beyond this popular MoE  setup, we now ablate these dimensions and explore their impact on on-device model quality, latency and memory. Specifically, we're exploring a suite of eight model ablations trained for 50,000 steps using a range of active and total parameter counts. Figure~\ref{fig:exp_ablation} summarizes our findings along the active expert (left) and total expert (right) dimensions. For the active expert ablations, we fix the number of total experts to be 8, while the total expert ablations are fixed along the active parameter count (active experts=2).

\paragraph{Active Expert Ablation:} A larger number of active experts and, hence, a larger number of forward FLOPs improves model quality. However, approaching 8 active experts, returns are diminishing. Looking at the generation speed between settings, we find that the generation speed decreases linearly, while the peak memory is constant across increasing numbers of active experts\footnote{The peak memory would increase between settings if we actively offload experts.}.

\paragraph{Total Expert Ablation:} In this setup, model quality increases near linearly with the number of total experts. However, in comparison to the active parameter ablation, the quality improvement is less prominent (compare scales between sub-graphs). In regards of the generation speed and peak memory, increasing the total expert count does not impact generation speed, since the active experts and, hence, FLOPs are fixed. However, the number of total experts significantly impacts the peak memory consumption\footnote{The peak memory would be constant if we actively offload experts, however, this would further impact the generation speed.}.

To summarize, increasing the number of active and total experts improves model quality, however, requires a trade-off regarding either generation speed (i.e. latency) or memory.

\begin{figure}[h!]
    \centering
    \begin{tikzpicture}
        \begin{axis}[
            width=0.7\linewidth,
            height=6cm,
            xlabel={Training Steps},
            ylabel={Avg. LM Evals},
            xtick=data,
            xticklabels={0, 10k, 20k, 30k, 40k, 50k, 100k, 150k, 200k, 250k, 300k},
            legend pos=south east,
            grid=major,
            grid style=dashed,
        ]
        \addplot[
            color=blue,
            mark=square,
        ] coordinates {
            (0, 32.94)
            (1, 44.31)
            (2, 46.07)
            (3, 47.46)
            (4, 47.64)
            (5, 48.18)
            (6, 48.6)
            (7, 49.31)
            (8, 49.27)
            (9, 49.57)
            (10, 49.54)
        };
        \addplot[
            color=red,
            mark=diamond,
        ] coordinates {
            (0, 32.94)
            (1, 44.65)
            (2, 49.02)
            (3, 50.41)
            (4, 50.41)
            (5, 51.46)
            (6, 51.82)
            (7, 52.52)
            (8, 52.49)
            (9, 52.9)
            (10, 52.8)
        };
        \addplot[
            color=black,
            mark=triangle,
        ] coordinates {
            (0, 32.94)
            (1, 45.52)
            (2, 47.66)
            (3, 49.15)
            (4, 49.84)
            (5, 50.19)
            (6, 50.67)
            (7, 51.54)
            (8, 51.7)
            (9, 52.18)
            (10, 52.22)
        };
        \legend{1.4B Dense, 3.6B Dense, 1.4B MoE + WD}
        \addplot[
            color=black,
            dotted,
            thick,
        ] coordinates {
            (0, 49.84)
            (1, 49.84)
            (2, 49.84)
            (3, 49.84)
            (4, 49.84)
            (5, 49.84)
            (6, 49.84)
            (7, 49.84)
            (8, 49.84)
            (9, 49.84)
            (10, 49.84)
        };
        \end{axis}
    \end{tikzpicture}
    \caption{Training dynamics across different model candidates}
    \label{fig:train_proc}
\end{figure}
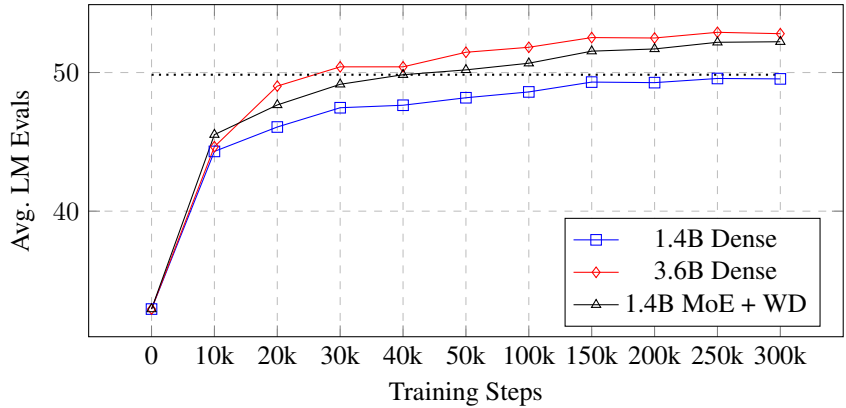

\section{Training Efficiency}
\label{app:train_eff}
In Figure \ref{fig:train_proc}, we're taking a look at the training process itself, comparing the training dynamic between MoE and dense model candidates, aligned by datasets, steps and hyper-parameters. Specifically, we compare the average language modeling performance between models at training checkpoints ranging from 10k to the full 310k steps. 

Comparing the active parameter and total parameter aligned dense models with our best performing MoE model, we corroborate the findings in \citet{lin2024momaefficientearlyfusionpretraining}, showing a 5-10x training efficiency gain using MoE models over their active parameter aligned dense candidates. Specifically, our MoE model candidate reaches the best performance of the 1.4B dense model at around 35k steps, while the larger and more powerful 3.6B dense model achieves generally higher scores.

\section{Related Work}
\paragraph{Small Scale Language Models}
With foundational models getting increasingly expensive to train and deploy, a dedicated effort has been made to develop small scale language models, aiming to enable foundational models to be deployed on-device (e.g. phones and glasses) or save compute during training and inference. Around those goals, two major research streams have formed: 

(1) Improving small-scale foundational model architectures. For example, the MobileLLM paper \cite{liu2024mobilellmoptimizingsubbillionparameter}, proposes deeper, narrower models at the sub-1B scale to perform better than shallower and wider networks. Similarly \citet{thawakar2024mobillamaaccuratelightweightfully} propose MobiLlama, showing that both, training and deployment cost can be reduced when using carefully curated parameter sharing schemes. Lastly, the BabyLlama series \cite{timiryasov2023babyllamaknowledgedistillation,tastet2024babyllama2ensembledistilledmodelsconsistently} shows that distilling knowledge from multiple teacher models leads to performance improvements under data-constrained conditions. 

(2) Improving the training data. Previous work along this line explicitly aims to improve model performance through cleaner, more streamlined data. For example, the Microsoft Phi series \cite{abdin2024phi3technicalreporthighly} shows that using curated textbook data for small language model pre-training can significantly improve model performance. Furthermore, Huggingface showed that highly curated, education-style data can greatly support the language modeling task for small language models in their SMoLLM \cite{benallal2024smollmcorpus} and fine-web \cite{lozhkov2024fineweb-edu} efforts.

For a more in-depth comparison of small language models, we refer readers to \citet{vannguyen2024surveysmalllanguagemodels}.

\paragraph{Sparse Mixture of Experts}
Mixture of Experts (MoEs) and, specifically, sparse Mixture of Experts have been exhaustively explored across different model sizes, including Qwen \citeyearpar{bai2023qwentechnicalreport,yang2024qwen2technicalreport} and OLMoE \citeyearpar{muennighoff2024olmoeopenmixtureofexpertslanguage} at the 1-3B active parameter scale, around 7B active parameters in the Mixtral \citeyearpar{jiang2024mixtralexperts} and Deepseek \citeyearpar{deepseekai2024deepseekv2strongeconomicalefficient} models, all the way up to DBRX \cite{introducing-dbrx} counting 36B and Grok-1 \cite{grok-os} with 86B active parameter. Exploring training and inference trends, as well as major design decisions, the OlMoE paper \cite{muennighoff2024olmoeopenmixtureofexpertslanguage} presents an important milestone in the development of MoE models, specifically at smaller scales. Here, we follow many of the OLMoE findings in our model selection and definition. Specifically comparing the inner workings of large MoE models in regards to the role of different experts, \citet{lo2024closerlookmixtureofexpertslarge} compare Mixtral \cite{jiang2024mixtralexperts}, Grok \cite{grok-os} and DeepSeek \cite{deepseekai2024deepseekv2strongeconomicalefficient} models, resulting in initial differences being found between model architectures, despite their different training paradigms. Here, we aim to make similar comnparisons, however, focus on fairness between models, removing as many confounding factors as possible during our model comparison. For a more detailed exploration of previous MoE settings, we refer readers to \citet{cai2024surveymixtureexperts}.

\paragraph{Weight Decomposition for Mixture of Experts}
As one of our extensions in this work, we propose a weight-decomposed version of a spare mixture of expert model. Along similar lines, \citet{dou2024loramoealleviateworldknowledge} previously proposed a low ranking (LoRa) style extension of dense networks, effectively turning them into a mixture of expert model during the supervised fine-tuning (SFT) stage. By freezing the dense backbone model and using a router in the SFT stage, the authors argue that the final model is more robust against catastrophic forgetting of the pre-training knowledge. In comparison to their approach, we apply the weight-decomposition in the pre-training stage, directly training the backbone model using more parameter-efficient experts.

\paragraph{Inference Efficiency}
Lastly, we explore more efficient MoE parameter offloading through the use of our novel BlES loss term, closely related to previous work to enhance model offloading during model inference. Specifically, \citet{xue2024moeinfinityoffloadingefficientmoemodel} present ``MoE-Infinity'', improving model expert pre-fetching and expert caching to reduce the number of model parameter transfers to and from the GPU. Similarly, EdgeMoE \cite{yi2023edgemoefastondeviceinference} presents an inference framework to enhance MoE offloading latency through predictive offloading and bitwidth adaptations. Furthermore, other inference optimization frameworks exist, such as Mixtral Fast Inference \cite{eliseev2023fastinferencemixtureofexpertslanguage} and DeepSpeed Efficient Inference \cite{aminabadi2022deepspeedinferenceenablingefficient}. Compared to this line of previous work, our approach is orthogonal, reducing the number of offloading actions during the model training stage, rather than at inference time.

\section{Conclusion}
In this work, we show how to enable sparse MoE architectures for on-device inference use-cases along the three on-device dimensions: Quality, Memory and Latency. 
Specifically, we show that in a fair comparison, MoE-style models outperform their dense counterparts on language modeling tasks by over over +2.35\%. Introducing our novel weight-decomposed experts, we show further performance gains of up to +1.1\% compared to standard MoE models. 
To truly enable MoE-style models for on-device use-cases, we tackle the model offloading bottleneck by reducing expert offloads in the training stage and, in turn, reduce model inference latency. Our ``grouped expert selection'' loss term thereby improves expert offloading efficiency by 6x and increases generation speed by 50\% compared to standard offloaded MoE models.

With the results presented in this paper, we effectively pave the way to unlock the potential of MoE-style architectures in on-device scenarios, supporting high quality, privacy preserving foundational models for edge devices. 

\bibliographystyle{assets/plainnat}
\bibliography{anthology,custom}

\begin{thebibliography}{39}
\providecommand{\natexlab}[1]{#1}
\providecommand{\url}[1]{\texttt{#1}}
\expandafter\ifx\csname urlstyle\endcsname\relax
  \providecommand{\doi}[1]{doi: #1}\else
  \providecommand{\doi}{doi: \begingroup \urlstyle{rm}\Url}\fi

\bibitem[Abdin et~al.(2024)Abdin, Aneja, Awadalla, Awadallah, Awan, Bach,
  Bahree, Bakhtiari, Bao, Behl, Benhaim, Bilenko, Bjorck, Bubeck, Cai, Cai,
  Chaudhary, Chen, Chen, Chen, Chen, Chen, Cheng, Chopra, Dai, Dixon, Eldan,
  Fragoso, Gao, Gao, Gao, Garg, Giorno, Goswami, Gunasekar, Haider, Hao,
  Hewett, Hu, Huynh, Iter, Jacobs, Javaheripi, Jin, Karampatziakis, Kauffmann,
  Khademi, Kim, Kim, Kurilenko, Lee, Lee, Li, Li, Liang, Liden, Lin, Lin, Liu,
  Liu, Liu, Liu, Liu, Luo, Madan, Mahmoudzadeh, Majercak, Mazzola, Mendes,
  Mitra, Modi, Nguyen, Norick, Patra, Perez-Becker, Portet, Pryzant, Qin,
  Radmilac, Ren, de~Rosa, Rosset, Roy, Ruwase, Saarikivi, Saied, Salim,
  Santacroce, Shah, Shang, Sharma, Shen, Shukla, Song, Tanaka, Tupini,
  Vaddamanu, Wang, Wang, Wang, Wang, Wang, Wang, Ward, Wen, Witte, Wu, Wu,
  Wyatt, Xiao, Xu, Xu, Xu, Xue, Yadav, Yang, Yang, Yang, Yang, Yu, Yuan, Zhang,
  Zhang, Zhang, Zhang, Zhang, Zhang, Zhang, and
  Zhou]{abdin2024phi3technicalreporthighly}
Marah Abdin, Jyoti Aneja, Hany Awadalla, Ahmed Awadallah, Ammar~Ahmad Awan,
  Nguyen Bach, Amit Bahree, Arash Bakhtiari, Jianmin Bao, Harkirat Behl, Alon
  Benhaim, Misha Bilenko, Johan Bjorck, Sébastien Bubeck, Martin Cai, Qin Cai,
  Vishrav Chaudhary, Dong Chen, Dongdong Chen, Weizhu Chen, Yen-Chun Chen,
  Yi-Ling Chen, Hao Cheng, Parul Chopra, Xiyang Dai, Matthew Dixon, Ronen
  Eldan, Victor Fragoso, Jianfeng Gao, Mei Gao, Min Gao, Amit Garg, Allie~Del
  Giorno, Abhishek Goswami, Suriya Gunasekar, Emman Haider, Junheng Hao,
  Russell~J. Hewett, Wenxiang Hu, Jamie Huynh, Dan Iter, Sam~Ade Jacobs, Mojan
  Javaheripi, Xin Jin, Nikos Karampatziakis, Piero Kauffmann, Mahoud Khademi,
  Dongwoo Kim, Young~Jin Kim, Lev Kurilenko, James~R. Lee, Yin~Tat Lee, Yuanzhi
  Li, Yunsheng Li, Chen Liang, Lars Liden, Xihui Lin, Zeqi Lin, Ce~Liu, Liyuan
  Liu, Mengchen Liu, Weishung Liu, Xiaodong Liu, Chong Luo, Piyush Madan, Ali
  Mahmoudzadeh, David Majercak, Matt Mazzola, Caio César~Teodoro Mendes,
  Arindam Mitra, Hardik Modi, Anh Nguyen, Brandon Norick, Barun Patra, Daniel
  Perez-Becker, Thomas Portet, Reid Pryzant, Heyang Qin, Marko Radmilac,
  Liliang Ren, Gustavo de~Rosa, Corby Rosset, Sambudha Roy, Olatunji Ruwase,
  Olli Saarikivi, Amin Saied, Adil Salim, Michael Santacroce, Shital Shah, Ning
  Shang, Hiteshi Sharma, Yelong Shen, Swadheen Shukla, Xia Song, Masahiro
  Tanaka, Andrea Tupini, Praneetha Vaddamanu, Chunyu Wang, Guanhua Wang, Lijuan
  Wang, Shuohang Wang, Xin Wang, Yu~Wang, Rachel Ward, Wen Wen, Philipp Witte,
  Haiping Wu, Xiaoxia Wu, Michael Wyatt, Bin Xiao, Can Xu, Jiahang Xu, Weijian
  Xu, Jilong Xue, Sonali Yadav, Fan Yang, Jianwei Yang, Yifan Yang, Ziyi Yang,
  Donghan Yu, Lu~Yuan, Chenruidong Zhang, Cyril Zhang, Jianwen Zhang, Li~Lyna
  Zhang, Yi~Zhang, Yue Zhang, Yunan Zhang, and Xiren Zhou.
\newblock Phi-3 technical report: A highly capable language model locally on
  your phone, 2024.
\newblock \url{https://arxiv.org/abs/2404.14219}.

\bibitem[Aminabadi et~al.(2022)Aminabadi, Rajbhandari, Zhang, Awan, Li, Li,
  Zheng, Rasley, Smith, Ruwase, and
  He]{aminabadi2022deepspeedinferenceenablingefficient}
Reza~Yazdani Aminabadi, Samyam Rajbhandari, Minjia Zhang, Ammar~Ahmad Awan,
  Cheng Li, Du~Li, Elton Zheng, Jeff Rasley, Shaden Smith, Olatunji Ruwase, and
  Yuxiong He.
\newblock Deepspeed inference: Enabling efficient inference of transformer
  models at unprecedented scale, 2022.
\newblock \url{https://arxiv.org/abs/2207.00032}.

\bibitem[Bai et~al.(2023)Bai, Bai, Chu, Cui, Dang, Deng, Fan, Ge, Han, Huang,
  Hui, Ji, Li, Lin, Lin, Liu, Liu, Lu, Lu, Ma, Men, Ren, Ren, Tan, Tan, Tu,
  Wang, Wang, Wang, Wu, Xu, Xu, Yang, Yang, Yang, Yang, Yao, Yu, Yuan, Yuan,
  Zhang, Zhang, Zhang, Zhang, Zhou, Zhou, Zhou, and
  Zhu]{bai2023qwentechnicalreport}
Jinze Bai, Shuai Bai, Yunfei Chu, Zeyu Cui, Kai Dang, Xiaodong Deng, Yang Fan,
  Wenbin Ge, Yu~Han, Fei Huang, Binyuan Hui, Luo Ji, Mei Li, Junyang Lin, Runji
  Lin, Dayiheng Liu, Gao Liu, Chengqiang Lu, Keming Lu, Jianxin Ma, Rui Men,
  Xingzhang Ren, Xuancheng Ren, Chuanqi Tan, Sinan Tan, Jianhong Tu, Peng Wang,
  Shijie Wang, Wei Wang, Shengguang Wu, Benfeng Xu, Jin Xu, An~Yang, Hao Yang,
  Jian Yang, Shusheng Yang, Yang Yao, Bowen Yu, Hongyi Yuan, Zheng Yuan,
  Jianwei Zhang, Xingxuan Zhang, Yichang Zhang, Zhenru Zhang, Chang Zhou,
  Jingren Zhou, Xiaohuan Zhou, and Tianhang Zhu.
\newblock Qwen technical report, 2023.
\newblock \url{https://arxiv.org/abs/2309.16609}.

\bibitem[Ben~Allal et~al.(2024)Ben~Allal, Lozhkov, Penedo, Wolf, and von
  Werra]{benallal2024smollmcorpus}
Loubna Ben~Allal, Anton Lozhkov, Guilherme Penedo, Thomas Wolf, and Leandro von
  Werra.
\newblock Smollm-corpus, 2024.
\newblock \url{https://huggingface.co/datasets/HuggingFaceTB/smollm-corpus}.

\bibitem[Cai et~al.(2024)Cai, Jiang, Wang, Tang, Kim, and
  Huang]{cai2024surveymixtureexperts}
Weilin Cai, Juyong Jiang, Fan Wang, Jing Tang, Sunghun Kim, and Jiayi Huang.
\newblock A survey on mixture of experts, 2024.
\newblock \url{https://arxiv.org/abs/2407.06204}.

\bibitem[Computer(2023)]{together2023redpajama}
Together Computer.
\newblock Redpajama: An open source recipe to reproduce llama training dataset,
  2023.
\newblock \url{https://github.com/togethercomputer/RedPajama-Data}.

\bibitem[{Databricks}(2023)]{introducing-dbrx}
{Databricks}.
\newblock Introducing dbrx: A new state-of-the-art open llm, 2023.
\newblock
  \url{https://www.databricks.com/blog/introducing-dbrx-new-state-art-open-llm}.
\newblock Blog post.

\bibitem[DeepSeek-AI(2024)]{deepseekai2024deepseekv2strongeconomicalefficient}
DeepSeek-AI.
\newblock Deepseek-v2: A strong, economical, and efficient mixture-of-experts
  language model, 2024.
\newblock \url{https://arxiv.org/abs/2405.04434}.

\bibitem[Dou et~al.(2024)Dou, Zhou, Liu, Gao, Zhao, Shen, Zhou, Xi, Wang, Fan,
  Pu, Zhu, Zheng, Gui, Zhang, and Huang]{dou2024loramoealleviateworldknowledge}
Shihan Dou, Enyu Zhou, Yan Liu, Songyang Gao, Jun Zhao, Wei Shen, Yuhao Zhou,
  Zhiheng Xi, Xiao Wang, Xiaoran Fan, Shiliang Pu, Jiang Zhu, Rui Zheng, Tao
  Gui, Qi~Zhang, and Xuanjing Huang.
\newblock Loramoe: Alleviate world knowledge forgetting in large language
  models via moe-style plugin, 2024.
\newblock \url{https://arxiv.org/abs/2312.09979}.

\bibitem[Eliseev and
  Mazur(2023)]{eliseev2023fastinferencemixtureofexpertslanguage}
Artyom Eliseev and Denis Mazur.
\newblock Fast inference of mixture-of-experts language models with offloading,
  2023.
\newblock \url{https://arxiv.org/abs/2312.17238}.

\bibitem[Fedus et~al.(2022)Fedus, Zoph, and
  Shazeer]{fedus2022switchtransformersscalingtrillion}
William Fedus, Barret Zoph, and Noam Shazeer.
\newblock Switch transformers: Scaling to trillion parameter models with simple
  and efficient sparsity, 2022.
\newblock \url{https://arxiv.org/abs/2101.03961}.

\bibitem[Gao et~al.(2024)Gao, Tow, Abbasi, Biderman, Black, DiPofi, Foster,
  Golding, Hsu, Le~Noac'h, Li, McDonell, Muennighoff, Ociepa, Phang, Reynolds,
  Schoelkopf, Skowron, Sutawika, Tang, Thite, Wang, Wang, and
  Zou]{eval-harness}
Leo Gao, Jonathan Tow, Baber Abbasi, Stella Biderman, Sid Black, Anthony
  DiPofi, Charles Foster, Laurence Golding, Jeffrey Hsu, Alain Le~Noac'h,
  Haonan Li, Kyle McDonell, Niklas Muennighoff, Chris Ociepa, Jason Phang,
  Laria Reynolds, Hailey Schoelkopf, Aviya Skowron, Lintang Sutawika, Eric
  Tang, Anish Thite, Ben Wang, Kevin Wang, and Andy Zou.
\newblock A framework for few-shot language model evaluation, 07 2024.
\newblock \url{https://zenodo.org/records/12608602}.

\bibitem[Grattafiori et~al.(2024)Grattafiori, Dubey, Jauhri, Pandey, Kadian,
  Al-Dahle, Letman, Mathur, Schelten, Vaughan, Yang, Fan, Goyal, Hartshorn,
  Yang, Mitra, Sravankumar, Korenev, Hinsvark, Rao, Zhang, Rodriguez,
  Gregerson, Spataru, Roziere, Biron, Tang, Chern, Caucheteux, Nayak, Bi,
  Marra, McConnell, Keller, Touret, Wu, Wong, Ferrer, Nikolaidis, Allonsius,
  Song, Pintz, Livshits, Wyatt, Esiobu, Choudhary, Mahajan, Garcia-Olano,
  Perino, Hupkes, Lakomkin, AlBadawy, Lobanova, Dinan, Smith, Radenovic,
  Guzmán, Zhang, Synnaeve, Lee, Anderson, Thattai, Nail, Mialon, Pang,
  Cucurell, Nguyen, Korevaar, Xu, Touvron, Zarov, Ibarra, Kloumann, Misra,
  Evtimov, Zhang, Copet, Lee, Geffert, Vranes, Park, Mahadeokar, Shah, van~der
  Linde, Billock, Hong, Lee, Fu, Chi, Huang, Liu, Wang, Yu, Bitton, Spisak,
  Park, Rocca, Johnstun, Saxe, Jia, Alwala, Prasad, Upasani, Plawiak, Li,
  Heafield, Stone, El-Arini, Iyer, Malik, Chiu, Bhalla, Lakhotia,
  Rantala-Yeary, van~der Maaten, Chen, Tan, Jenkins, Martin, Madaan, Malo,
  Blecher, Landzaat, de~Oliveira, Muzzi, Pasupuleti, Singh, Paluri, Kardas,
  Tsimpoukelli, Oldham, Rita, Pavlova, Kambadur, Lewis, Si, Singh, Hassan,
  Goyal, Torabi, Bashlykov, Bogoychev, Chatterji, Zhang, Duchenne, Çelebi,
  Alrassy, Zhang, Li, Vasic, Weng, Bhargava, Dubal, Krishnan, Koura, Xu, He,
  Dong, Srinivasan, Ganapathy, Calderer, Cabral, Stojnic, Raileanu, Maheswari,
  Girdhar, Patel, Sauvestre, Polidoro, Sumbaly, Taylor, Silva, Hou, Wang,
  Hosseini, Chennabasappa, Singh, Bell, Kim, Edunov, Nie, Narang, Raparthy,
  Shen, Wan, Bhosale, Zhang, Vandenhende, Batra, Whitman, Sootla, Collot,
  Gururangan, Borodinsky, Herman, Fowler, Sheasha, Georgiou, Scialom,
  Speckbacher, Mihaylov, Xiao, Karn, Goswami, Gupta, Ramanathan, Kerkez,
  Gonguet, Do, Vogeti, Albiero, Petrovic, Chu, Xiong, Fu, Meers, Martinet,
  Wang, Wang, Tan, Xia, Xie, Jia, Wang, Goldschlag, Gaur, Babaei, Wen, Song,
  Zhang, Li, Mao, Coudert, Yan, Chen, Papakipos, Singh, Srivastava, Jain,
  Kelsey, Shajnfeld, Gangidi, Victoria, Goldstand, Menon, Sharma, Boesenberg,
  Baevski, Feinstein, Kallet, Sangani, Teo, Yunus, Lupu, Alvarado, Caples, Gu,
  Ho, Poulton, Ryan, Ramchandani, Dong, Franco, Goyal, Saraf, Chowdhury,
  Gabriel, Bharambe, Eisenman, Yazdan, James, Maurer, Leonhardi, Huang, Loyd,
  Paola, Paranjape, Liu, Wu, Ni, Hancock, Wasti, Spence, Stojkovic, Gamido,
  Montalvo, Parker, Burton, Mejia, Liu, Wang, Kim, Zhou, Hu, Chu, Cai, Tindal,
  Feichtenhofer, Gao, Civin, Beaty, Kreymer, Li, Adkins, Xu, Testuggine, David,
  Parikh, Liskovich, Foss, Wang, Le, Holland, Dowling, Jamil, Montgomery,
  Presani, Hahn, Wood, Le, Brinkman, Arcaute, Dunbar, Smothers, Sun, Kreuk,
  Tian, Kokkinos, Ozgenel, Caggioni, Kanayet, Seide, Florez, Schwarz, Badeer,
  Swee, Halpern, Herman, Sizov, Guangyi, Zhang, Lakshminarayanan, Inan,
  Shojanazeri, Zou, Wang, Zha, Habeeb, Rudolph, Suk, Aspegren, Goldman, Zhan,
  Damlaj, Molybog, Tufanov, Leontiadis, Veliche, Gat, Weissman, Geboski, Kohli,
  Lam, Asher, Gaya, Marcus, Tang, Chan, Zhen, Reizenstein, Teboul, Zhong, Jin,
  Yang, Cummings, Carvill, Shepard, McPhie, Torres, Ginsburg, Wang, Wu, U,
  Saxena, Khandelwal, Zand, Matosich, Veeraraghavan, Michelena, Li, Jagadeesh,
  Huang, Chawla, Huang, Chen, Garg, A, Silva, Bell, Zhang, Guo, Yu, Moshkovich,
  Wehrstedt, Khabsa, Avalani, Bhatt, Mankus, Hasson, Lennie, Reso, Groshev,
  Naumov, Lathi, Keneally, Liu, Seltzer, Valko, Restrepo, Patel, Vyatskov,
  Samvelyan, Clark, Macey, Wang, Hermoso, Metanat, Rastegari, Bansal,
  Santhanam, Parks, White, Bawa, Singhal, Egebo, Usunier, Mehta, Laptev, Dong,
  Cheng, Chernoguz, Hart, Salpekar, Kalinli, Kent, Parekh, Saab, Balaji,
  Rittner, Bontrager, Roux, Dollar, Zvyagina, Ratanchandani, Yuvraj, Liang,
  Alao, Rodriguez, Ayub, Murthy, Nayani, Mitra, Parthasarathy, Li, Hogan,
  Battey, Wang, Howes, Rinott, Mehta, Siby, Bondu, Datta, Chugh, Hunt, Dhillon,
  Sidorov, Pan, Mahajan, Verma, Yamamoto, Ramaswamy, Lindsay, Lindsay, Feng,
  Lin, Zha, Patil, Shankar, Zhang, Zhang, Wang, Agarwal, Sajuyigbe, Chintala,
  Max, Chen, Kehoe, Satterfield, Govindaprasad, Gupta, Deng, Cho, Virk,
  Subramanian, Choudhury, Goldman, Remez, Glaser, Best, Koehler, Robinson, Li,
  Zhang, Matthews, Chou, Shaked, Vontimitta, Ajayi, Montanez, Mohan, Kumar,
  Mangla, Ionescu, Poenaru, Mihailescu, Ivanov, Li, Wang, Jiang, Bouaziz,
  Constable, Tang, Wu, Wang, Wu, Gao, Kleinman, Chen, Hu, Jia, Qi, Li, Zhang,
  Zhang, Adi, Nam, Yu, Wang, Zhao, Hao, Qian, Li, He, Rait, DeVito, Rosnbrick,
  Wen, Yang, Zhao, and Ma]{grattafiori2024llama3herdmodels}
Aaron Grattafiori, Abhimanyu Dubey, Abhinav Jauhri, Abhinav Pandey, Abhishek
  Kadian, Ahmad Al-Dahle, Aiesha Letman, Akhil Mathur, Alan Schelten, Alex
  Vaughan, Amy Yang, Angela Fan, Anirudh Goyal, Anthony Hartshorn, Aobo Yang,
  Archi Mitra, Archie Sravankumar, Artem Korenev, Arthur Hinsvark, Arun Rao,
  Aston Zhang, Aurelien Rodriguez, Austen Gregerson, Ava Spataru, Baptiste
  Roziere, Bethany Biron, Binh Tang, Bobbie Chern, Charlotte Caucheteux, Chaya
  Nayak, Chloe Bi, Chris Marra, Chris McConnell, Christian Keller, Christophe
  Touret, Chunyang Wu, Corinne Wong, Cristian~Canton Ferrer, Cyrus Nikolaidis,
  Damien Allonsius, Daniel Song, Danielle Pintz, Danny Livshits, Danny Wyatt,
  David Esiobu, Dhruv Choudhary, Dhruv Mahajan, Diego Garcia-Olano, Diego
  Perino, Dieuwke Hupkes, Egor Lakomkin, Ehab AlBadawy, Elina Lobanova, Emily
  Dinan, Eric~Michael Smith, Filip Radenovic, Francisco Guzmán, Frank Zhang,
  Gabriel Synnaeve, Gabrielle Lee, Georgia~Lewis Anderson, Govind Thattai,
  Graeme Nail, Gregoire Mialon, Guan Pang, Guillem Cucurell, Hailey Nguyen,
  Hannah Korevaar, Hu~Xu, Hugo Touvron, Iliyan Zarov, Imanol~Arrieta Ibarra,
  Isabel Kloumann, Ishan Misra, Ivan Evtimov, Jack Zhang, Jade Copet, Jaewon
  Lee, Jan Geffert, Jana Vranes, Jason Park, Jay Mahadeokar, Jeet Shah, Jelmer
  van~der Linde, Jennifer Billock, Jenny Hong, Jenya Lee, Jeremy Fu, Jianfeng
  Chi, Jianyu Huang, Jiawen Liu, Jie Wang, Jiecao Yu, Joanna Bitton, Joe
  Spisak, Jongsoo Park, Joseph Rocca, Joshua Johnstun, Joshua Saxe, Junteng
  Jia, Kalyan~Vasuden Alwala, Karthik Prasad, Kartikeya Upasani, Kate Plawiak,
  Ke~Li, Kenneth Heafield, Kevin Stone, Khalid El-Arini, Krithika Iyer, Kshitiz
  Malik, Kuenley Chiu, Kunal Bhalla, Kushal Lakhotia, Lauren Rantala-Yeary,
  Laurens van~der Maaten, Lawrence Chen, Liang Tan, Liz Jenkins, Louis Martin,
  Lovish Madaan, Lubo Malo, Lukas Blecher, Lukas Landzaat, Luke de~Oliveira,
  Madeline Muzzi, Mahesh Pasupuleti, Mannat Singh, Manohar Paluri, Marcin
  Kardas, Maria Tsimpoukelli, Mathew Oldham, Mathieu Rita, Maya Pavlova,
  Melanie Kambadur, Mike Lewis, Min Si, Mitesh~Kumar Singh, Mona Hassan, Naman
  Goyal, Narjes Torabi, Nikolay Bashlykov, Nikolay Bogoychev, Niladri
  Chatterji, Ning Zhang, Olivier Duchenne, Onur Çelebi, Patrick Alrassy,
  Pengchuan Zhang, Pengwei Li, Petar Vasic, Peter Weng, Prajjwal Bhargava,
  Pratik Dubal, Praveen Krishnan, Punit~Singh Koura, Puxin Xu, Qing He,
  Qingxiao Dong, Ragavan Srinivasan, Raj Ganapathy, Ramon Calderer,
  Ricardo~Silveira Cabral, Robert Stojnic, Roberta Raileanu, Rohan Maheswari,
  Rohit Girdhar, Rohit Patel, Romain Sauvestre, Ronnie Polidoro, Roshan
  Sumbaly, Ross Taylor, Ruan Silva, Rui Hou, Rui Wang, Saghar Hosseini, Sahana
  Chennabasappa, Sanjay Singh, Sean Bell, Seohyun~Sonia Kim, Sergey Edunov,
  Shaoliang Nie, Sharan Narang, Sharath Raparthy, Sheng Shen, Shengye Wan,
  Shruti Bhosale, Shun Zhang, Simon Vandenhende, Soumya Batra, Spencer Whitman,
  Sten Sootla, Stephane Collot, Suchin Gururangan, Sydney Borodinsky, Tamar
  Herman, Tara Fowler, Tarek Sheasha, Thomas Georgiou, Thomas Scialom, Tobias
  Speckbacher, Todor Mihaylov, Tong Xiao, Ujjwal Karn, Vedanuj Goswami, Vibhor
  Gupta, Vignesh Ramanathan, Viktor Kerkez, Vincent Gonguet, Virginie Do, Vish
  Vogeti, Vítor Albiero, Vladan Petrovic, Weiwei Chu, Wenhan Xiong, Wenyin Fu,
  Whitney Meers, Xavier Martinet, Xiaodong Wang, Xiaofang Wang, Xiaoqing~Ellen
  Tan, Xide Xia, Xinfeng Xie, Xuchao Jia, Xuewei Wang, Yaelle Goldschlag,
  Yashesh Gaur, Yasmine Babaei, Yi~Wen, Yiwen Song, Yuchen Zhang, Yue Li,
  Yuning Mao, Zacharie~Delpierre Coudert, Zheng Yan, Zhengxing Chen, Zoe
  Papakipos, Aaditya Singh, Aayushi Srivastava, Abha Jain, Adam Kelsey, Adam
  Shajnfeld, Adithya Gangidi, Adolfo Victoria, Ahuva Goldstand, Ajay Menon,
  Ajay Sharma, Alex Boesenberg, Alexei Baevski, Allie Feinstein, Amanda Kallet,
  Amit Sangani, Amos Teo, Anam Yunus, Andrei Lupu, Andres Alvarado, Andrew
  Caples, Andrew Gu, Andrew Ho, Andrew Poulton, Andrew Ryan, Ankit Ramchandani,
  Annie Dong, Annie Franco, Anuj Goyal, Aparajita Saraf, Arkabandhu Chowdhury,
  Ashley Gabriel, Ashwin Bharambe, Assaf Eisenman, Azadeh Yazdan, Beau James,
  Ben Maurer, Benjamin Leonhardi, Bernie Huang, Beth Loyd, Beto~De Paola,
  Bhargavi Paranjape, Bing Liu, Bo~Wu, Boyu Ni, Braden Hancock, Bram Wasti,
  Brandon Spence, Brani Stojkovic, Brian Gamido, Britt Montalvo, Carl Parker,
  Carly Burton, Catalina Mejia, Ce~Liu, Changhan Wang, Changkyu Kim, Chao Zhou,
  Chester Hu, Ching-Hsiang Chu, Chris Cai, Chris Tindal, Christoph
  Feichtenhofer, Cynthia Gao, Damon Civin, Dana Beaty, Daniel Kreymer, Daniel
  Li, David Adkins, David Xu, Davide Testuggine, Delia David, Devi Parikh,
  Diana Liskovich, Didem Foss, Dingkang Wang, Duc Le, Dustin Holland, Edward
  Dowling, Eissa Jamil, Elaine Montgomery, Eleonora Presani, Emily Hahn, Emily
  Wood, Eric-Tuan Le, Erik Brinkman, Esteban Arcaute, Evan Dunbar, Evan
  Smothers, Fei Sun, Felix Kreuk, Feng Tian, Filippos Kokkinos, Firat Ozgenel,
  Francesco Caggioni, Frank Kanayet, Frank Seide, Gabriela~Medina Florez,
  Gabriella Schwarz, Gada Badeer, Georgia Swee, Gil Halpern, Grant Herman,
  Grigory Sizov, Guangyi, Zhang, Guna Lakshminarayanan, Hakan Inan, Hamid
  Shojanazeri, Han Zou, Hannah Wang, Hanwen Zha, Haroun Habeeb, Harrison
  Rudolph, Helen Suk, Henry Aspegren, Hunter Goldman, Hongyuan Zhan, Ibrahim
  Damlaj, Igor Molybog, Igor Tufanov, Ilias Leontiadis, Irina-Elena Veliche,
  Itai Gat, Jake Weissman, James Geboski, James Kohli, Janice Lam, Japhet
  Asher, Jean-Baptiste Gaya, Jeff Marcus, Jeff Tang, Jennifer Chan, Jenny Zhen,
  Jeremy Reizenstein, Jeremy Teboul, Jessica Zhong, Jian Jin, Jingyi Yang, Joe
  Cummings, Jon Carvill, Jon Shepard, Jonathan McPhie, Jonathan Torres, Josh
  Ginsburg, Junjie Wang, Kai Wu, Kam~Hou U, Karan Saxena, Kartikay Khandelwal,
  Katayoun Zand, Kathy Matosich, Kaushik Veeraraghavan, Kelly Michelena, Keqian
  Li, Kiran Jagadeesh, Kun Huang, Kunal Chawla, Kyle Huang, Lailin Chen,
  Lakshya Garg, Lavender A, Leandro Silva, Lee Bell, Lei Zhang, Liangpeng Guo,
  Licheng Yu, Liron Moshkovich, Luca Wehrstedt, Madian Khabsa, Manav Avalani,
  Manish Bhatt, Martynas Mankus, Matan Hasson, Matthew Lennie, Matthias Reso,
  Maxim Groshev, Maxim Naumov, Maya Lathi, Meghan Keneally, Miao Liu,
  Michael~L. Seltzer, Michal Valko, Michelle Restrepo, Mihir Patel, Mik
  Vyatskov, Mikayel Samvelyan, Mike Clark, Mike Macey, Mike Wang, Miquel~Jubert
  Hermoso, Mo~Metanat, Mohammad Rastegari, Munish Bansal, Nandhini Santhanam,
  Natascha Parks, Natasha White, Navyata Bawa, Nayan Singhal, Nick Egebo,
  Nicolas Usunier, Nikhil Mehta, Nikolay~Pavlovich Laptev, Ning Dong, Norman
  Cheng, Oleg Chernoguz, Olivia Hart, Omkar Salpekar, Ozlem Kalinli, Parkin
  Kent, Parth Parekh, Paul Saab, Pavan Balaji, Pedro Rittner, Philip Bontrager,
  Pierre Roux, Piotr Dollar, Polina Zvyagina, Prashant Ratanchandani, Pritish
  Yuvraj, Qian Liang, Rachad Alao, Rachel Rodriguez, Rafi Ayub, Raghotham
  Murthy, Raghu Nayani, Rahul Mitra, Rangaprabhu Parthasarathy, Raymond Li,
  Rebekkah Hogan, Robin Battey, Rocky Wang, Russ Howes, Ruty Rinott, Sachin
  Mehta, Sachin Siby, Sai~Jayesh Bondu, Samyak Datta, Sara Chugh, Sara Hunt,
  Sargun Dhillon, Sasha Sidorov, Satadru Pan, Saurabh Mahajan, Saurabh Verma,
  Seiji Yamamoto, Sharadh Ramaswamy, Shaun Lindsay, Shaun Lindsay, Sheng Feng,
  Shenghao Lin, Shengxin~Cindy Zha, Shishir Patil, Shiva Shankar, Shuqiang
  Zhang, Shuqiang Zhang, Sinong Wang, Sneha Agarwal, Soji Sajuyigbe, Soumith
  Chintala, Stephanie Max, Stephen Chen, Steve Kehoe, Steve Satterfield,
  Sudarshan Govindaprasad, Sumit Gupta, Summer Deng, Sungmin Cho, Sunny Virk,
  Suraj Subramanian, Sy~Choudhury, Sydney Goldman, Tal Remez, Tamar Glaser,
  Tamara Best, Thilo Koehler, Thomas Robinson, Tianhe Li, Tianjun Zhang, Tim
  Matthews, Timothy Chou, Tzook Shaked, Varun Vontimitta, Victoria Ajayi,
  Victoria Montanez, Vijai Mohan, Vinay~Satish Kumar, Vishal Mangla, Vlad
  Ionescu, Vlad Poenaru, Vlad~Tiberiu Mihailescu, Vladimir Ivanov, Wei Li,
  Wenchen Wang, Wenwen Jiang, Wes Bouaziz, Will Constable, Xiaocheng Tang,
  Xiaojian Wu, Xiaolan Wang, Xilun Wu, Xinbo Gao, Yaniv Kleinman, Yanjun Chen,
  Ye~Hu, Ye~Jia, Ye~Qi, Yenda Li, Yilin Zhang, Ying Zhang, Yossi Adi, Youngjin
  Nam, Yu, Wang, Yu~Zhao, Yuchen Hao, Yundi Qian, Yunlu Li, Yuzi He, Zach Rait,
  Zachary DeVito, Zef Rosnbrick, Zhaoduo Wen, Zhenyu Yang, Zhiwei Zhao, and
  Zhiyu Ma.
\newblock The llama 3 herd of models, 2024.
\newblock \url{https://arxiv.org/abs/2407.21783}.

\bibitem[Hu et~al.(2021)Hu, Shen, Wallis, Allen-Zhu, Li, Wang, Wang, and
  Chen]{hu2021loralowrankadaptationlarge}
Edward~J. Hu, Yelong Shen, Phillip Wallis, Zeyuan Allen-Zhu, Yuanzhi Li, Shean
  Wang, Lu~Wang, and Weizhu Chen.
\newblock Lora: Low-rank adaptation of large language models, 2021.
\newblock \url{https://arxiv.org/abs/2106.09685}.

\bibitem[Jacobs et~al.(1991)Jacobs, Jordan, Nowlan, and Hinton]{6797059}
Robert~A. Jacobs, Michael~I. Jordan, Steven~J. Nowlan, and Geoffrey~E. Hinton.
\newblock Adaptive mixtures of local experts.
\newblock \emph{Neural Computation}, 3\penalty0 (1):\penalty0 79--87, 1991.
\newblock \doi{10.1162/neco.1991.3.1.79}.

\bibitem[Jiang et~al.(2024)Jiang, Sablayrolles, Roux, Mensch, Savary, Bamford,
  Chaplot, de~las Casas, Hanna, Bressand, Lengyel, Bour, Lample, Lavaud,
  Saulnier, Lachaux, Stock, Subramanian, Yang, Antoniak, Scao, Gervet, Lavril,
  Wang, Lacroix, and Sayed]{jiang2024mixtralexperts}
Albert~Q. Jiang, Alexandre Sablayrolles, Antoine Roux, Arthur Mensch, Blanche
  Savary, Chris Bamford, Devendra~Singh Chaplot, Diego de~las Casas, Emma~Bou
  Hanna, Florian Bressand, Gianna Lengyel, Guillaume Bour, Guillaume Lample,
  Lélio~Renard Lavaud, Lucile Saulnier, Marie-Anne Lachaux, Pierre Stock,
  Sandeep Subramanian, Sophia Yang, Szymon Antoniak, Teven~Le Scao, Théophile
  Gervet, Thibaut Lavril, Thomas Wang, Timothée Lacroix, and William~El Sayed.
\newblock Mixtral of experts, 2024.
\newblock \url{https://arxiv.org/abs/2401.04088}.

\bibitem[Jordan and Jacobs(1993)]{716791}
M.I. Jordan and R.A. Jacobs.
\newblock Hierarchical mixtures of experts and the em algorithm.
\newblock In \emph{Proceedings of 1993 International Conference on Neural
  Networks (IJCNN-93-Nagoya, Japan)}, volume~2, pages 1339--1344 vol.2, 1993.
\newblock \doi{10.1109/IJCNN.1993.716791}.

\bibitem[Lepikhin et~al.(2020)Lepikhin, Lee, Xu, Chen, Firat, Huang, Krikun,
  Shazeer, and Chen]{lepikhin2020gshardscalinggiantmodels}
Dmitry Lepikhin, HyoukJoong Lee, Yuanzhong Xu, Dehao Chen, Orhan Firat, Yanping
  Huang, Maxim Krikun, Noam Shazeer, and Zhifeng Chen.
\newblock Gshard: Scaling giant models with conditional computation and
  automatic sharding, 2020.
\newblock \url{https://arxiv.org/abs/2006.16668}.

\bibitem[Lin et~al.(2024)Lin, Shrivastava, Luo, Iyer, Lewis, Ghosh,
  Zettlemoyer, and Aghajanyan]{lin2024momaefficientearlyfusionpretraining}
Xi~Victoria Lin, Akshat Shrivastava, Liang Luo, Srinivasan Iyer, Mike Lewis,
  Gargi Ghosh, Luke Zettlemoyer, and Armen Aghajanyan.
\newblock Moma: Efficient early-fusion pre-training with mixture of
  modality-aware experts, 2024.
\newblock \url{https://arxiv.org/abs/2407.21770}.

\bibitem[Liu et~al.(2024)Liu, Zhao, Iandola, Lai, Tian, Fedorov, Xiong, Chang,
  Shi, Krishnamoorthi, Lai, and
  Chandra]{liu2024mobilellmoptimizingsubbillionparameter}
Zechun Liu, Changsheng Zhao, Forrest Iandola, Chen Lai, Yuandong Tian, Igor
  Fedorov, Yunyang Xiong, Ernie Chang, Yangyang Shi, Raghuraman Krishnamoorthi,
  Liangzhen Lai, and Vikas Chandra.
\newblock Mobilellm: Optimizing sub-billion parameter language models for
  on-device use cases, 2024.
\newblock \url{https://arxiv.org/abs/2402.14905}.

\bibitem[Lo et~al.(2024)Lo, Huang, Qiu, Wang, and
  Fu]{lo2024closerlookmixtureofexpertslarge}
Ka~Man Lo, Zeyu Huang, Zihan Qiu, Zili Wang, and Jie Fu.
\newblock A closer look into mixture-of-experts in large language models, 2024.
\newblock \url{https://arxiv.org/abs/2406.18219}.

\bibitem[Lozhkov et~al.(2024)Lozhkov, Ben~Allal, von Werra, and
  Wolf]{lozhkov2024fineweb-edu}
Anton Lozhkov, Loubna Ben~Allal, Leandro von Werra, and Thomas Wolf.
\newblock Fineweb-edu: the finest collection of educational content, 2024.
\newblock \url{https://huggingface.co/datasets/HuggingFaceFW/fineweb-edu}.

\bibitem[Muennighoff et~al.(2024)Muennighoff, Soldaini, Groeneveld, Lo,
  Morrison, Min, Shi, Walsh, Tafjord, Lambert, Gu, Arora, Bhagia, Schwenk,
  Wadden, Wettig, Hui, Dettmers, Kiela, Farhadi, Smith, Koh, Singh, and
  Hajishirzi]{muennighoff2024olmoeopenmixtureofexpertslanguage}
Niklas Muennighoff, Luca Soldaini, Dirk Groeneveld, Kyle Lo, Jacob Morrison,
  Sewon Min, Weijia Shi, Pete Walsh, Oyvind Tafjord, Nathan Lambert, Yuling Gu,
  Shane Arora, Akshita Bhagia, Dustin Schwenk, David Wadden, Alexander Wettig,
  Binyuan Hui, Tim Dettmers, Douwe Kiela, Ali Farhadi, Noah~A. Smith, Pang~Wei
  Koh, Amanpreet Singh, and Hannaneh Hajishirzi.
\newblock Olmoe: Open mixture-of-experts language models, 2024.
\newblock \url{https://arxiv.org/abs/2409.02060}.

\bibitem[Nguyen et~al.(2024)Nguyen, Shen, Aponte, Xia, Basu, Hu, Chen, Parmar,
  Kunapuli, Barrow, Wu, Singh, Wang, Gu, Dernoncourt, Ahmed, Lipka, Zhang,
  Chen, Yu, Kim, Deilamsalehy, Park, Rimer, Zhang, Yang, Rossi, and
  Nguyen]{vannguyen2024surveysmalllanguagemodels}
Chien~Van Nguyen, Xuan Shen, Ryan Aponte, Yu~Xia, Samyadeep Basu, Zhengmian Hu,
  Jian Chen, Mihir Parmar, Sasidhar Kunapuli, Joe Barrow, Junda Wu, Ashish
  Singh, Yu~Wang, Jiuxiang Gu, Franck Dernoncourt, Nesreen~K. Ahmed, Nedim
  Lipka, Ruiyi Zhang, Xiang Chen, Tong Yu, Sungchul Kim, Hanieh Deilamsalehy,
  Namyong Park, Mike Rimer, Zhehao Zhang, Huanrui Yang, Ryan~A. Rossi, and
  Thien~Huu Nguyen.
\newblock A survey of small language models, 2024.
\newblock \url{https://arxiv.org/abs/2410.20011}.

\bibitem[Penedo et~al.(2024)Penedo, Kydlíček, allal, Lozhkov, Mitchell,
  Raffel, Werra, and Wolf]{penedo2024finewebdatasetsdecantingweb}
Guilherme Penedo, Hynek Kydlíček, Loubna~Ben allal, Anton Lozhkov, Margaret
  Mitchell, Colin Raffel, Leandro~Von Werra, and Thomas Wolf.
\newblock The fineweb datasets: Decanting the web for the finest text data at
  scale, 2024.
\newblock \url{https://arxiv.org/abs/2406.17557}.

\bibitem[{PyTorch Labs}(2023)]{gpt-fast}
{PyTorch Labs}.
\newblock Gpt-fast, 2023.
\newblock \url{https://github.com/pytorch-labs/gpt-fast}.
\newblock GitHub repository.

\bibitem[Raffel et~al.(2020)Raffel, Shazeer, Roberts, Lee, Narang, Matena,
  Zhou, Li, and Liu]{JMLR:v21:20-074}
Colin Raffel, Noam Shazeer, Adam Roberts, Katherine Lee, Sharan Narang, Michael
  Matena, Yanqi Zhou, Wei Li, and Peter~J. Liu.
\newblock Exploring the limits of transfer learning with a unified text-to-text
  transformer.
\newblock \emph{Journal of Machine Learning Research}, 21\penalty0
  (140):\penalty0 1--67, 2020.
\newblock \url{http://jmlr.org/papers/v21/20-074.html}.

\bibitem[Shazeer et~al.(2017)Shazeer, Mirhoseini, Maziarz, Davis, Le, Hinton,
  and Dean]{shazeer2017outrageouslylargeneuralnetworks}
Noam Shazeer, Azalia Mirhoseini, Krzysztof Maziarz, Andy Davis, Quoc Le,
  Geoffrey Hinton, and Jeff Dean.
\newblock Outrageously large neural networks: The sparsely-gated
  mixture-of-experts layer, 2017.
\newblock \url{https://arxiv.org/abs/1701.06538}.

\bibitem[Tastet and
  Timiryasov(2024)]{tastet2024babyllama2ensembledistilledmodelsconsistently}
Jean-Loup Tastet and Inar Timiryasov.
\newblock Babyllama-2: Ensemble-distilled models consistently outperform
  teachers with limited data, 2024.
\newblock \url{https://arxiv.org/abs/2409.17312}.

\bibitem[Thawakar et~al.(2024)Thawakar, Vayani, Khan, Cholakal, Anwer,
  Felsberg, Baldwin, Xing, and
  Khan]{thawakar2024mobillamaaccuratelightweightfully}
Omkar Thawakar, Ashmal Vayani, Salman Khan, Hisham Cholakal, Rao~M. Anwer,
  Michael Felsberg, Tim Baldwin, Eric~P. Xing, and Fahad~Shahbaz Khan.
\newblock Mobillama: Towards accurate and lightweight fully transparent gpt,
  2024.
\newblock \url{https://arxiv.org/abs/2402.16840}.

\bibitem[Timiryasov and
  Tastet(2023)]{timiryasov2023babyllamaknowledgedistillation}
Inar Timiryasov and Jean-Loup Tastet.
\newblock Baby llama: knowledge distillation from an ensemble of teachers
  trained on a small dataset with no performance penalty, 2023.
\newblock \url{https://arxiv.org/abs/2308.02019}.

\bibitem[Vaswani et~al.(2023)Vaswani, Shazeer, Parmar, Uszkoreit, Jones, Gomez,
  Kaiser, and Polosukhin]{vaswani2023attentionneed}
Ashish Vaswani, Noam Shazeer, Niki Parmar, Jakob Uszkoreit, Llion Jones,
  Aidan~N. Gomez, Lukasz Kaiser, and Illia Polosukhin.
\newblock Attention is all you need, 2023.
\newblock \url{https://arxiv.org/abs/1706.03762}.

\bibitem[Weber et~al.(2024)Weber, Fu, Anthony, Oren, Adams, Alexandrov, Lyu,
  Nguyen, Yao, Adams, Athiwaratkun, Chalamala, Chen, Ryabinin, Dao, Liang, Ré,
  Rish, and Zhang]{weber2024redpajama}
Maurice Weber, Daniel~Y. Fu, Quentin Anthony, Yonatan Oren, Shane Adams, Anton
  Alexandrov, Xiaozhong Lyu, Huu Nguyen, Xiaozhe Yao, Virginia Adams, Ben
  Athiwaratkun, Rahul Chalamala, Kezhen Chen, Max Ryabinin, Tri Dao, Percy
  Liang, Christopher Ré, Irina Rish, and Ce~Zhang.
\newblock Redpajama: an open dataset for training large language models.
\newblock \emph{NeurIPS Datasets and Benchmarks Track}, 2024.

\bibitem[Wolf et~al.(2020)Wolf, Debut, Sanh, Chaumond, Delangue, Moi, Cistac,
  Rault, Louf, Funtowicz, Davison, Shleifer, von Platen, Ma, Jernite, Plu, Xu,
  Scao, Gugger, Drame, Lhoest, and
  Rush]{wolf2020huggingfacestransformersstateoftheartnatural}
Thomas Wolf, Lysandre Debut, Victor Sanh, Julien Chaumond, Clement Delangue,
  Anthony Moi, Pierric Cistac, Tim Rault, Rémi Louf, Morgan Funtowicz, Joe
  Davison, Sam Shleifer, Patrick von Platen, Clara Ma, Yacine Jernite, Julien
  Plu, Canwen Xu, Teven~Le Scao, Sylvain Gugger, Mariama Drame, Quentin Lhoest,
  and Alexander~M. Rush.
\newblock Huggingface's transformers: State-of-the-art natural language
  processing, 2020.
\newblock \url{https://arxiv.org/abs/1910.03771}.

\bibitem[{x.ai}(2023)]{grok-os}
{x.ai}.
\newblock Grok os, 2023.
\newblock \url{https://x.ai/blog/grok-os}.
\newblock Blog post.

\bibitem[Xue et~al.(2024)Xue, Fu, Lu, Mai, and
  Marina]{xue2024moeinfinityoffloadingefficientmoemodel}
Leyang Xue, Yao Fu, Zhan Lu, Luo Mai, and Mahesh Marina.
\newblock Moe-infinity: Offloading-efficient moe model serving, 2024.
\newblock \url{https://arxiv.org/abs/2401.14361}.

\bibitem[Yang et~al.(2024)Yang, Yang, Hui, Zheng, Yu, Zhou, Li, Li, Liu, Huang,
  Dong, Wei, Lin, Tang, Wang, Yang, Tu, Zhang, Ma, Yang, Xu, Zhou, Bai, He,
  Lin, Dang, Lu, Chen, Yang, Li, Xue, Ni, Zhang, Wang, Peng, Men, Gao, Lin,
  Wang, Bai, Tan, Zhu, Li, Liu, Ge, Deng, Zhou, Ren, Zhang, Wei, Ren, Liu, Fan,
  Yao, Zhang, Wan, Chu, Liu, Cui, Zhang, Guo, and
  Fan]{yang2024qwen2technicalreport}
An~Yang, Baosong Yang, Binyuan Hui, Bo~Zheng, Bowen Yu, Chang Zhou, Chengpeng
  Li, Chengyuan Li, Dayiheng Liu, Fei Huang, Guanting Dong, Haoran Wei, Huan
  Lin, Jialong Tang, Jialin Wang, Jian Yang, Jianhong Tu, Jianwei Zhang,
  Jianxin Ma, Jianxin Yang, Jin Xu, Jingren Zhou, Jinze Bai, Jinzheng He,
  Junyang Lin, Kai Dang, Keming Lu, Keqin Chen, Kexin Yang, Mei Li, Mingfeng
  Xue, Na~Ni, Pei Zhang, Peng Wang, Ru~Peng, Rui Men, Ruize Gao, Runji Lin,
  Shijie Wang, Shuai Bai, Sinan Tan, Tianhang Zhu, Tianhao Li, Tianyu Liu,
  Wenbin Ge, Xiaodong Deng, Xiaohuan Zhou, Xingzhang Ren, Xinyu Zhang, Xipin
  Wei, Xuancheng Ren, Xuejing Liu, Yang Fan, Yang Yao, Yichang Zhang, Yu~Wan,
  Yunfei Chu, Yuqiong Liu, Zeyu Cui, Zhenru Zhang, Zhifang Guo, and Zhihao Fan.
\newblock Qwen2 technical report, 2024.
\newblock \url{https://arxiv.org/abs/2407.10671}.

\bibitem[Yi et~al.(2023)Yi, Guo, Wei, Zhou, Wang, and
  Xu]{yi2023edgemoefastondeviceinference}
Rongjie Yi, Liwei Guo, Shiyun Wei, Ao~Zhou, Shangguang Wang, and Mengwei Xu.
\newblock Edgemoe: Fast on-device inference of moe-based large language models,
  2023.
\newblock \url{https://arxiv.org/abs/2308.14352}.

\bibitem[Zhou et~al.(2022)Zhou, Lei, Liu, Du, Huang, Zhao, Dai, Chen, Le, and
  Laudon]{zhou2022mixtureofexpertsexpertchoicerouting}
Yanqi Zhou, Tao Lei, Hanxiao Liu, Nan Du, Yanping Huang, Vincent Zhao, Andrew
  Dai, Zhifeng Chen, Quoc Le, and James Laudon.
\newblock Mixture-of-experts with expert choice routing, 2022.
\newblock \url{https://arxiv.org/abs/2202.09368}.

\end{thebibliography}
\end{document}